\documentclass[11pt]{article}

\usepackage[]{acl}

\usepackage{times}
\usepackage{array}
\usepackage{latexsym}
\usepackage{booktabs}
\usepackage{graphicx}

\usepackage{amsmath}
\usepackage{amssymb}
\usepackage{amstext}

\usepackage[T1]{fontenc}
\usepackage[utf8]{inputenc}

\usepackage{courier}  
\usepackage{xcolor}
\usepackage{listings}

\usepackage{enumitem}
\setlist{nosep}

\lstdefinestyle{codeblock}{
  basicstyle=\ttfamily\scriptsize,
  breaklines=true,
  breakatwhitespace=true,
  columns=fullflexible,
  keepspaces=true,
  frame=single,
  framerule=0.3pt,
  rulecolor=\color{black!40},
  xleftmargin=0.5em,
  xrightmargin=0.5em,
  linewidth=\columnwidth,
  showstringspaces=false,
  tabsize=2
}
\lstdefinelanguage{json}{
  morestring=[b]",
  morecomment=[l]{//},
  morekeywords={true,false,null},
  sensitive=false
}
\usepackage{tikz}
\usepackage{placeins}

\usepackage{wasysym}

\title{Beyond Memorization: Testing LLM Reasoning on Unseen Theory of Computation Tasks}

\author{Shlok Shelat \\ Ahmedabad University\\ Gujarat, India \\  
       \And  
Jay Raval \\ Ahmedabad University \\ Gujarat, India \\      
       \And
Souvik Roy \\ Ahmedabad University \\ Gujarat, India  \\   
       \And
Manas Gaur \\ University of Maryland \\ Baltimore County \\ Baltimore, MD, USA 
}

\begin{document}
\maketitle

\begin{abstract}
Large language models (LLMs) have demonstrated strong performance on formal language tasks, yet whether this reflects genuine symbolic reasoning or pattern matching on familiar constructions remains unclear. We introduce a benchmark for deterministic finite automata (DFA) construction from regular languages, comprising factual knowledge questions, seen construction problems from public sources, and two types of unseen problems: hand-crafted instances with multiple interacting constraints and systematically generated problems via Arden's theorem. Models achieve perfect accuracy on factual questions and 84-90\% on seen tasks. However, accuracy drops sharply on unseen problems (by 30-64\%), with failures stemming from systematic misinterpretation of language constraints, incorrect handling of Kleene-star semantics, and a failure to preserve global consistency. We evaluate a three-stage hint protocol that enables correction of shallow errors but does not reliably resolve globally inconsistent or structurally flawed automata. Our analysis across multiple prompting strategies (direct, Chain-of-Thought, Tree-of-Thought) reveals that errors persist regardless of prompting approach, exposing a fundamental gap between LLMs' ability to generate syntactically plausible DFAs and their capacity for semantically correct formal reasoning.
\end{abstract}

\section{Introduction}

Large language models (LLMs) have demonstrated remarkable performance on diverse reasoning benchmarks, from mathematical problem-solving~\cite{10.5555/3600270.3600548,welleck2021,azerbayev2023} to code generation~\cite{10.5555/3600270.3602614}. However, a fundamental question remains unresolved: do these models perform genuine symbolic reasoning, or do they primarily rely on pattern matching over memorized examples? Recent work reveals persistent failures on tasks requiring structured symbolic manipulation~\cite{katz2025,yue2024}, suggesting that strong benchmark performance may not reflect robust reasoning capabilities.

We address this question through the lens of \emph{deterministic finite automata (DFA) construction from regular languages}, a core problem in the Theory of Computation (ToC). Moreover, this also depicts the task of lexical analyser which are solved by using tools like Lex, Flex with limitations~\cite{10.5555/1177220}. This task offers unique advantages as a reasoning probe: (1) correctness is formally verifiable through exhaustive testing, (2) solutions require multi-step symbolic manipulation with global consistency constraints, (3) the space of possible problem instances is combinatorially vast, and (4) DFAs represent the simplest non-trivial computational model, ensuring that failures cannot be attributed to problem complexity or ambiguous specifications. We focus on prompting-based evaluation (rather than fine-tuning) to reflect practical LLM usage. Critically, while existing ToC benchmarks~\cite{10.1145/3649165.3690116,zahraei2024} evaluate factual knowledge and proof verification, they do not systematically control for \emph{memorization versus compositional generalization}, a model may succeed by recalling similar problems from training data rather than reasoning from first principles.

To isolate genuine reasoning capability, we introduce a carefully designed benchmark with three components: (1) a \emph{knowledge-checking dataset} verifying mastery of DFA definitions and properties, (2) a \emph{seen construction dataset} comprising 90 publicly available DFA problems, and (3) an \emph{unseen construction dataset} with 180 novel problems generated via two complementary approaches. The first approach, \emph{mathematical art}, manually constructs problems with multiple interacting constraints, forbidden substrings, and narrative-based specifications (e.g., encoding chess moves). The second approach, \emph{mathematical engineering}, systematically generates problems via Arden's theorem~\cite{sipser13}, producing structurally complex regular expressions unlikely to appear in training data. This seen/unseen split enables us to measure the performance gap attributable to memorization versus reasoning.

We evaluate frontier LLMs: GPT-5.1, Gemini-2.5-Flash, and Grok-4.1-fast-reasoning -- across multiple prompting strategies including Chain-of-Thought (CoT) and Tree-of-Thought (ToT)~\cite{wang2024c,10.5555/3666122.3666639}. We also introduce a \emph{three-stage hint protocol} that progressively reveals construction errors, enabling us to assess whether LLMs can self-correct when guided.

\noindent\textbf{Main Findings.} Our results reveal a stark dissociation between knowledge and reasoning: all models achieve 100\% accuracy on factual questions and 84--90\% on seen construction tasks, but accuracy drops sharply on unseen problems (20.67--59.12\% under direct prompting, representing 30--64 \% point drops). Detailed error analysis shows systematic failure modes: incorrect simplification of Kleene star semantics, failure to preserve constraints under concatenation, and introduction of spurious states. Critically, these failures persist across all prompting strategies, and the hint protocol primarily corrects shallow errors while leaving globally inconsistent automata uncorrected.

\noindent\textbf{Contributions.} This work makes the following contributions: (i) \textbf{Novel benchmark:} We introduce the first DFA construction benchmark with systematic seen/unseen splits, comprising 50 knowledge, 90 seen, and 180 unseen problems (60 hand-crafted, 120 via Arden's theorem). (ii) \textbf{ Controlled memorization study:} By evaluating structurally similar seen/unseen pairs, we provide the first evidence that LLM success on ToC tasks primarily reflects memorization rather than compositional reasoning. (iii) \textbf{ Comprehensive prompting evaluation:} We evaluate CoT, ToT (with four construction methods: direct, minimization, derivative-based, Thompson's algorithm), and a novel hint-based self-correction protocol. (iv) \textbf{Systematic failure taxonomy:} Through detailed analysis of 500+ incorrect DFAs, we identify six recurring failure modes (derivative normalization errors, constraint composition failures, etc.) that reveal fundamental limitations in symbolic state tracking. All datasets, prompts, evaluation code, and model outputs are released at \url{https://anonymous.4open.science/r/dfa-llm-evaluation-B82D/} to support reproducibility and future research on formal reasoning in LLMs.

\section{Seen Dataset and Task Formulation}
\subsection{Task Definition}
We evaluate LLMs on the task of constructing deterministic finite automata (DFAs) from formal language specifications. Given a target language $L$ specified either as a regular expression (RE) or natural language description, the model must produce a DFA $D = \langle \mathcal{Q}, \Sigma, \delta, q_0, F \rangle$ that recognizes exactly $L$. Here, $\mathcal{Q}$ is a finite set of states, $\Sigma$ is the input alphabet, $\delta : \mathcal{Q} \times \Sigma \rightarrow \mathcal{Q}$ is the transition function, $q_0 \in \mathcal{Q}$ is the start state, and $F \subseteq \mathcal{Q}$ is the set of accepting states~\cite{sipser13}. A DFA $D$ accepts string $w = w_1 w_2 \cdots w_n \in \Sigma^*$ if there exists a sequence of states $s_0, s_1, \ldots, s_n$ such that (i) $s_0 = q_0$, (ii) $s_i = \delta(s_{i-1}, w_i)$ for all $1 \leq i \leq n$, and (iii) $s_n \in F$. We say $D$ recognizes language $L$ if $L = \{w \in \Sigma^* \mid D \text{ accepts } w\}$. A language is \emph{regular} if there exists a DFA that recognizes it. Correct DFA construction requires models to: (1) parse and interpret formal language specifications, (2) identify the minimal state structure capturing all constraints, (3) design transitions ensuring acceptance of all and only valid strings, and (4) maintain global consistency across all states and symbols. Critically, there exist exponentially many invalid DFAs for any language, and small errors in state semantics or transition assignments can invalidate the entire construction.

\begin{figure*}
\begin{center}
\begin{tabular}{ccc}
\includegraphics[width=1.6in]{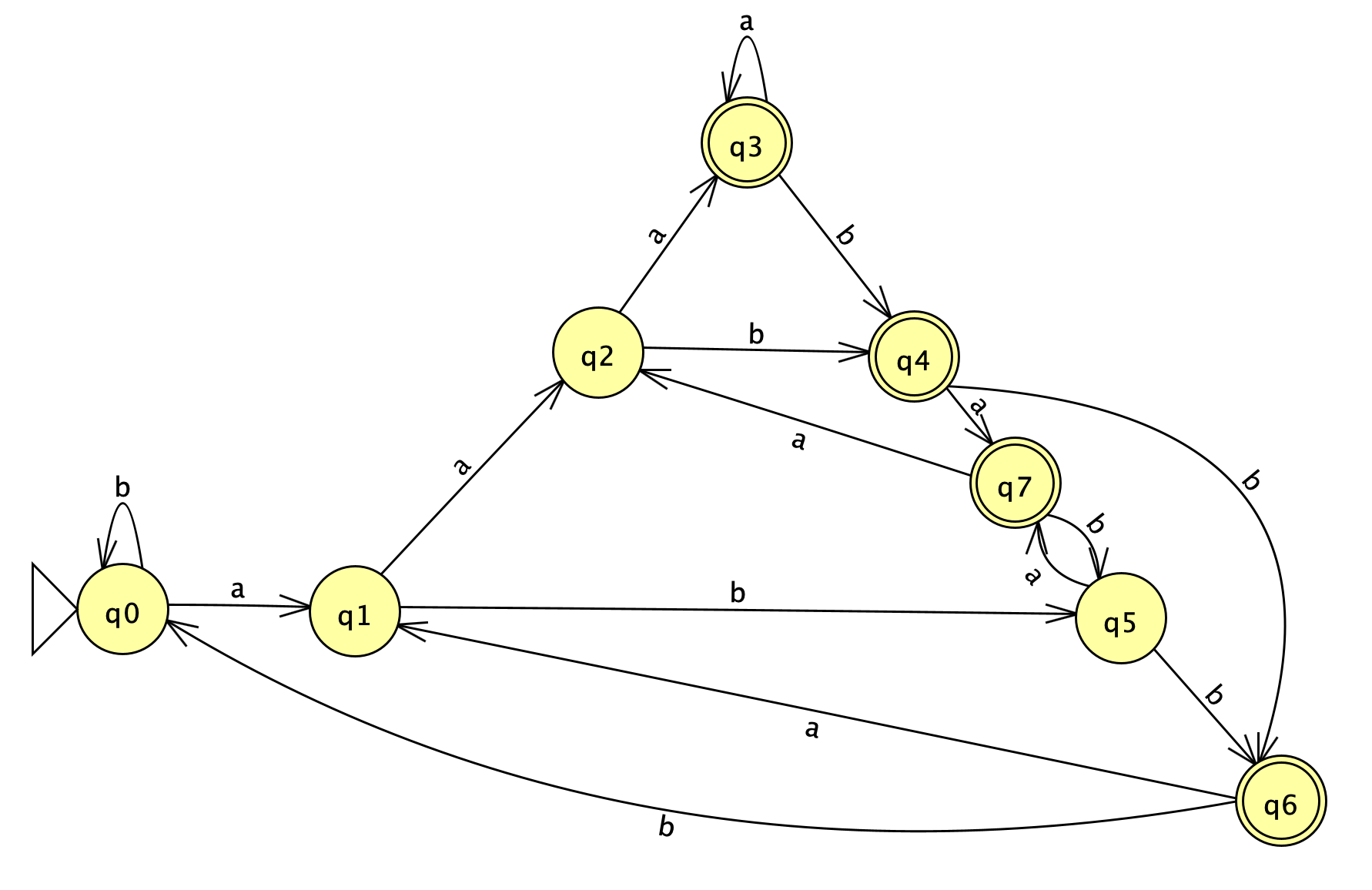} & \includegraphics[width=2.2in]{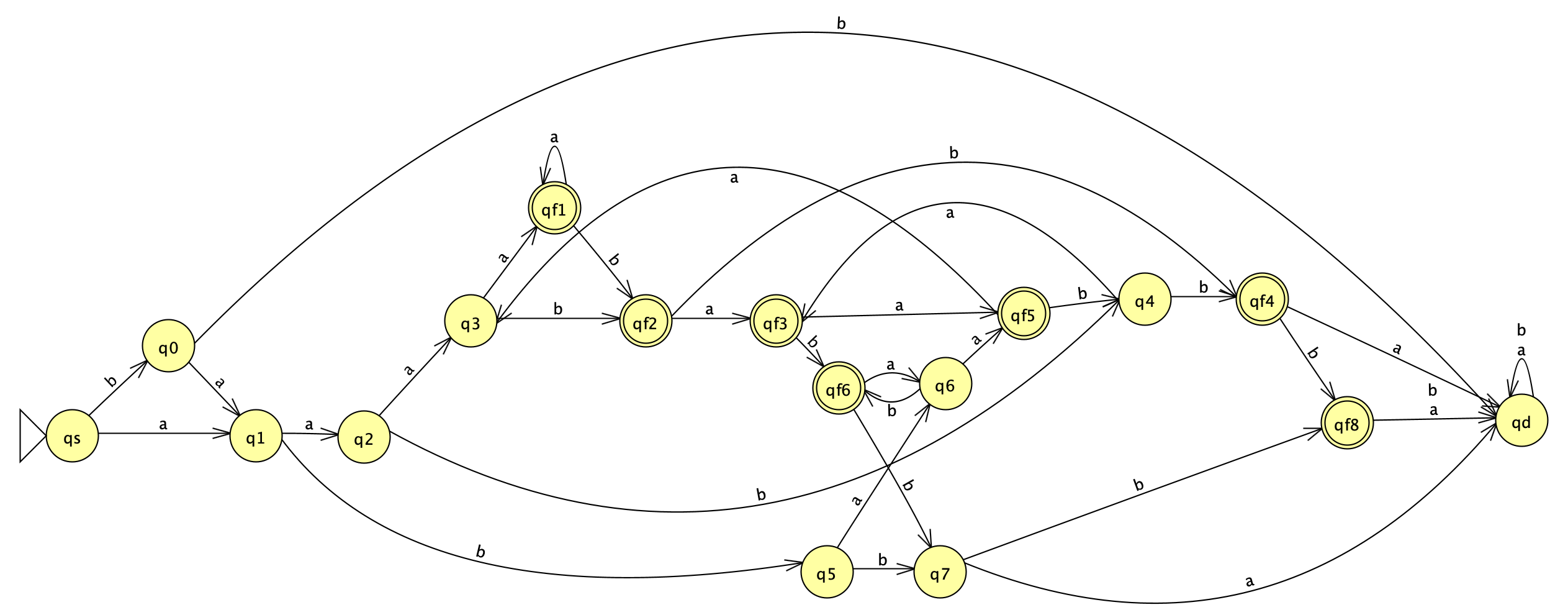} & \includegraphics[width=1.7in]{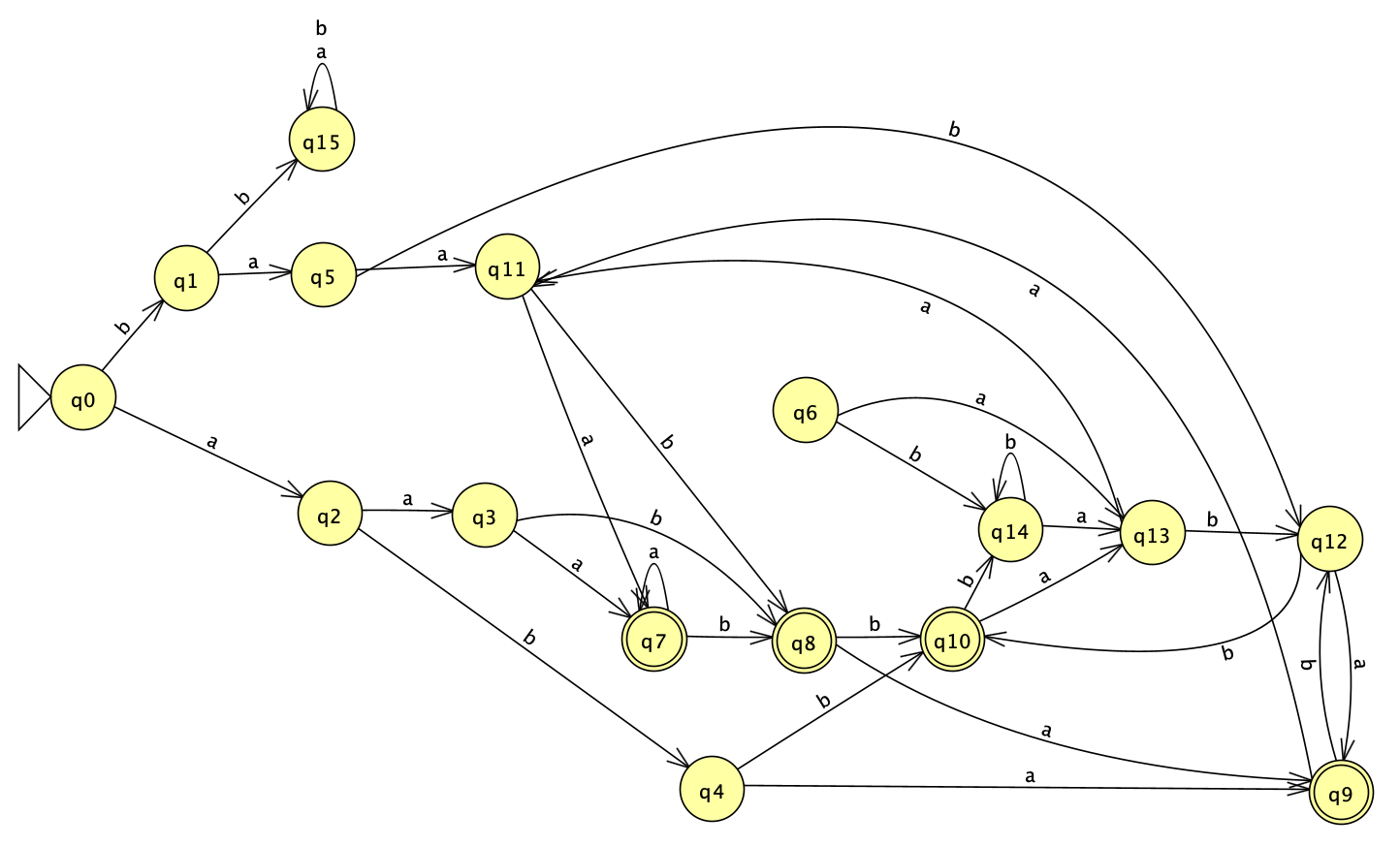} \\
(a) Correct minimal DFA for $L_1$ & (b) Correct minimal DFA for $L_2$ & (c) Gemini-2.5 output for $L_2$\\
\end{tabular}
\caption{\small{(a): Minimum DFA for language $L_1$; (b) \& (c): Comparison of correct minimal DFA for unseen language $L_2$ (see b) and representative incorrect output (see c). Notation: $\bigcirc$ = state, $\triangleright$ = start state, $\circledcirc$ = accepting state. Transition $q_0 \xrightarrow{a} q_1$ indicates that on input $a$ from state $q_0$, the DFA transitions to state $q_1${\color{red} (Please zoom for better readability)}.}}
\label{fig:L2}
\end{center}
\end{figure*}

\noindent\textbf{Problem Instances.} Each problem specifies the target language in one of two formats: \textit{(a) Natural language:} $L_1$ = \{Construct a DFA over $\{a,b\}$ that accepts all strings in which the third-to-last symbol from the end must be `a'\}. Here, if the problem format is given in natural language, it is not possible to design DFA using Lex, Flex~\cite{10.5555/1177220}. \textit{(b) RE:} $L_1 = (a+b)^*a(a+b)(a+b)$. Both formats specify the same language (Figure~\ref{fig:L2}(a)). Models must output a complete DFA specification including states, transitions, start state, and accepting states. We focus on \emph{minimal} DFAs (fewest states) when possible, as minimality requires identifying semantically equivalent states -- a reasoning challenge beyond simply enumerating valid transitions. However, during checking the performance of LLMs, we consider only correct DFA (particularly, not minimal DFA).

\subsection{Knowledge and Seen DFA Dataset}
\label{sec:seen_results}

Before evaluating construction ability, we verify that models possess foundational knowledge of automata theory. This dataset comprises 50 questions: 25 multiple-choice and 25 short-answer questions covering DFA definitions, RE properties, and the relationship between DFAs and nondeterministic FAs (NFAs). All questions are drawn from standard ToC textbooks and publicly available materials. \textbf{Results:} All models achieve 100\% accuracy (Table~\ref{tab:seen_results}), confirming complete mastery of relevant definitions and properties. Consequently, errors on construction tasks cannot be attributed to missing conceptual knowledge.

The seen dataset comprises 90 DFA construction problems collected from online university problem sets, textbooks, and publicly accessible ToC resources. For each problem, we verified that both the problem statement and solution DFA are publicly available online, ensuring these represent patterns likely encountered during pretraining. 

\noindent\textbf{Language Characteristics.} The seen dataset includes standard patterns such as: (i) Suffix/prefix constraints (e.g., ``ends with $ab$''); (ii) Counting modulo $k$ (e.g., ``even number of $a$'s''); (iii) Position-based constraints (e.g., ``$3^{rd}$-last symbol $a$''); and (iv) Boolean combinations (e.g., ``contains $aa$ or $bb$'').

\begin{table}[t]
\centering
\scriptsize
\begin{tabular}{lcccc}
\toprule
\textbf{Dataset} & \textbf{Prompting} & \textbf{GPT-5.1} & \textbf{Grok-4.1} & \textbf{Gemini-2.5} \\
\midrule
Knowledge & Zero-Shot & 100\% & 100\% & 100\% \\
Seen DFA & Zero-Shot & 84.2\% & 89.5\% & 85.0\% \\
\bottomrule
\end{tabular}
\caption{\small{Results of knowledge and seen DFA datasets.}}
\label{tab:seen_results}
\end{table}

\noindent\textbf{Evaluation Protocol.} We evaluate models using zero-shot direct prompting (no worked examples). Models are queried via official APIs with temperature set to $0$ for deterministic decoding. Each model receives the problem specification and must output a complete DFA in structured JSON format specifying states, alphabet, transitions, start state, and accepting states (Appendix~\ref{app:prompts}).

\noindent\textbf{Results.} Table~\ref{tab:seen_results} shows that all models achieve strong performance: Grok-4.1-fast-reasoning achieves the highest success rate (89.5\%), followed by Gemini-2.5-Flash (85.0\%) and GPT-5.1 (84.2\%). For the example language $L_1$, all three models produce the correct minimal DFA. These results suggest that models can successfully construct DFAs for \emph{familiar} problem patterns.

\noindent\textbf{The Memorization Question.} High accuracy on seen problems does not imply genuine reasoning capability. Models may succeed by retrieving similar examples from training data rather than performing compositional symbolic manipulation. To isolate reasoning from memorization, we next introduce the unseen construction dataset with carefully controlled novelty.

\subsection{The Memorization Question}
\label{sec:problem_formulation}

Strong performance on the seen dataset (84--90\% accuracy) might suggest that LLMs possess robust DFA construction capabilities. However, this conclusion is premature: models may succeed by retrieving memorized solution patterns rather than performing genuine symbolic reasoning. To distinguish these explanations, we must evaluate performance on \emph{structurally novel} problems that require compositional generalization. \textbf{Motivating Example:} Consider two closely related languages: \textbf{$L_1$ (seen):} Accepts all strings over $\{a,b\}$ where the third-to-last symbol is `a'; \textbf{$L_2$ (unseen):} Accepts all strings over $\{a,b\}$ where (i) the fourth-to-last symbol is `a', \emph{and} (ii) substring `bb' does not appear before any `a'. Language $L_2$ extends $L_1$ with one additional constraint (forbidding `bb' before `a') and a minor modification to the position constraint ($4^{th}$-last instead of $3^{rd}$-last). Both languages require similar reasoning -- tracking symbol positions while enforcing ordering constraints -- yet $L_2$ is constructed to be absent from public problem sets and textbook solutions.

\noindent\textbf{Empirical Evidence.} Despite achieving 100\% accuracy on $L_1$, \emph{all models fail on $L_2$ under direct prompting}. Figure~\ref{fig:L2}(c) shows Gemini-2.5-Flash's output: while the constructed DFA accepts some valid strings (e.g.,~`aaaa',~`aaab'), it also accepts invalid strings such as `aaa' (violates position constraint), and `aabbaba' (violates both constraints). Similar systematic errors occur for GPT-5.1 and Grok-4.1-fast-reasoning (Appendix~\ref{app:dfa_outputs}). This failure is particularly revealing because $L_1$ and $L_2$ differ only in \emph{constraint composition}, not in fundamental reasoning requirements. The models correctly construct DFAs for simpler positional constraints (seen in training) but fail when these constraints are combined with ordering restrictions -- suggesting success on seen tasks reflects pattern retrieval rather than robust symbolic reasoning.

\noindent\textbf{Research Question.} These observations motivate our central research question: \emph{To what extent does LLM performance on formal reasoning tasks depend on memorization of training examples versus compositional symbolic reasoning?} Answering this question requires systematic evaluation on unseen problems that control for structural novelty while preserving task and reasoning requirements.

\noindent\textbf{Unseen Dataset Design.} To enable this controlled evaluation, we construct an unseen DFA dataset using two complementary approaches: (i) \textbf{Mathematical Art (60 problems):} Manually designed problems with multiple interacting constraints, forbidden substrings, positional patterns, and narrative-based specifications (e.g., encoding chess openings, maze navigation). These problems require creative constraint combination absent from standard curricula. (ii) \textbf{Mathematical Engineering (120 problems):} Systematically generated problems via Arden's theorem~\cite{sipser13}. We construct random NFAs, derive their accepted languages through algebraic elimination, and use the resulting regular expressions (often highly nested and non-standard) as problem specifications. This approach ensures structural diversity and scalability.

All unseen problems are manually verified to be absent from public problem repositories, textbooks, and online course materials. We categorize problems by difficulty (easy, medium, hard) based on the number of constraints (Art) or nesting depth (Engineering), enabling fine-grained analysis of where reasoning breaks down. Detailed construction procedures and representative examples are provided in Section~\ref{sec:unseen_dataset}.

\noindent\textbf{Evaluation Framework.} Beyond measuring accuracy on unseen problems, we evaluate multiple prompting strategies (CoT, ToT with four construction branches) and introduce a three-stage hint protocol to assess self-correction capability. This comprehensive evaluation isolates whether failures stem from initial misinterpretation, inability to maintain symbolic consistency, or fundamental reasoning deficits that persist even with corrective guidance.

\section{Unseen DFA Construction Dataset}
\label{sec:unseen_dataset}
We employ two generation strategies: \emph{manual constraint composition} (60 problems) for creative problem design, and \emph{systematic Arden's theorem inversion} (120 problems) for scalable generation with guaranteed structural diversity.
\subsection{Manual Constraint Composition}
This approach extends seen DFA patterns through controlled increases in constraint complexity. Following the design principles illustrated by the $L_1 \rightarrow L_2$ transformation (Section~\ref{sec:problem_formulation}), we systematically combine multiple constraints that rarely co-occur in standard curricula: (i) \textbf{Product constructions:} Multiple conjunctive/disjunctive constraints requiring state-space products (e.g., ``strings starting with `aba' OR `bab' AND ending with their reverse''). (ii) \textbf{Interacting constraints:} Independent conditions that cannot be verified locally (e.g., ``fourth-to-last symbol is `a' AND substring `bb' never precedes `a''' -- language $L_2$). (iii) \textbf{Structural restrictions:} Positional patterns and forbidden substrings (e.g., ``XOR of first three bits equals final bit''; ``every 4-bit window's product is divisible by 4''). (iv) \textbf{Narrative encodings:} Real-world scenarios requiring formalization (e.g., chess opening sequences encoded as moves; fruit-mixing recipes as syrup combinations; see Appendix~\ref{app:datasets} for additional examples). All 60 problems were manually verified to be absent from the top 100 Google search results, standard textbooks \cite{sipser13, Hopcroft01}, and popular online repositories ({\tt GitHub}, {\tt StackOverflow}, course websites). This approach yields high-quality problems testing creative constraint integration, but scalability is limited by human effort.
\subsection{Generation via Arden's Theorem}
To address scalability while ensuring structural novelty, we employ a reverse-engineering approach: generate random NFAs, derive their accepted languages algebraically using Arden's theorem, and use the resulting regular expressions as problem specifications. This systematic procedure guarantees that each generated problem has a unique structure determined by random NFA topology rather than memorized patterns. Following is the generation procedure: (i) \textbf{Random NFA construction:} Random NFA with $5-8$ states, $2-3$ accepting states, and random transition density $0.3-0.6$. The nondeterministic transition function $\delta: \mathcal{Q} \times \Sigma \rightarrow 2^{\mathcal{Q}}$ permits multiple successor states per input symbol. (ii) \textbf{Language derivation via Arden's theorem:} Apply Arden's theorem~\cite{sipser13} -- if regular expressions $P$ and $Q$ satisfy $R = Q + RP$ and $P$ does not contain $\epsilon$, then $R = QP^*$ -- to eliminate states iteratively and derive a closed-form RE. (iii) \textbf{Minimal DFA construction:} Convert the NFA to a DFA via subset construction to obtain the ground-truth solution.
\begin{figure}[h!tbp]
\centering
\begin{tabular}{cc}
\includegraphics[width=1.5in, angle=90]{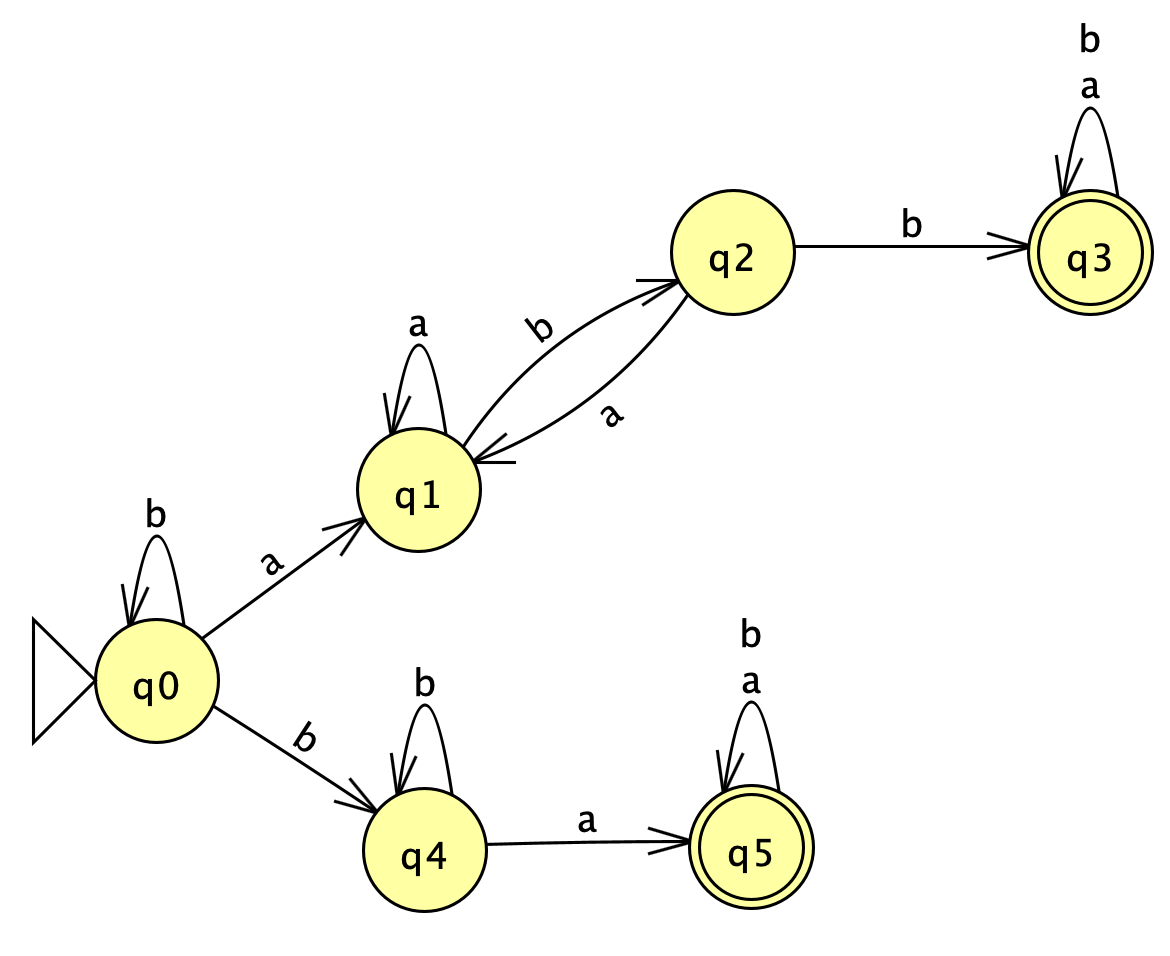} & \includegraphics[width=1.5in, angle=90]{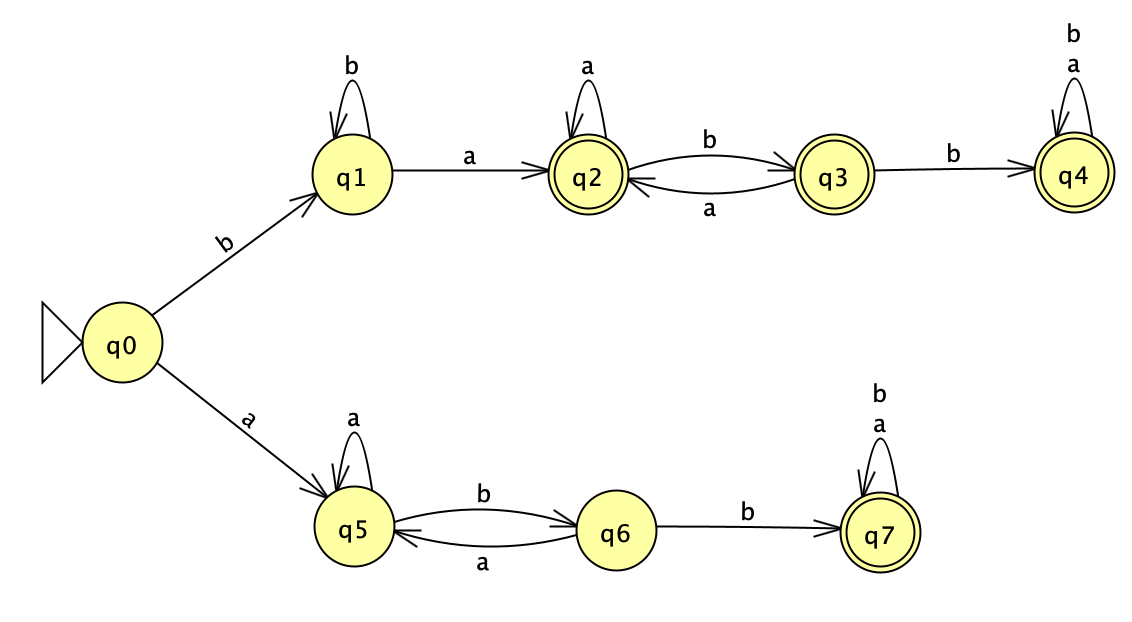} \\
(a)  & (b)
\end{tabular}
\caption{\textbf{(a):} Random NFA over $\{a,b\}$. \textbf{(b):} Minimal DFA recognizing the derived language $L_3$.}
\label{fig:L3}
\end{figure}

\noindent\textbf{Detailed Example: Deriving Language $L_3$.} We now demonstrate the complete derivation process using the random NFA shown in Figure~\ref{fig:L3}(a). 

\noindent\textbf{Step 1: State equations from NFA.} Each state's language is expressed recursively based on incoming transitions of Figure~\ref{fig:L3}(a):
\vspace{-0.3cm}
\begin{align}
q_5 &= q_5 a + q_5 b + q_4 a \label{eq:q5} \\
q_4 &= q_0 b + q_4 b \label{eq:q4} \\
q_3 &= q_3 a + q_3 b + q_2 b \label{eq:q3} \\
q_2 &= q_1 b \label{eq:q2} \\
q_1 &= q_0 a + q_1 a + q_2 a \label{eq:q1} \\
q_0 &= \epsilon + q_0 b \label{eq:q0}
\end{align}

Here $q_0$ is the start state (accepting $\epsilon$), and $q_3$, $q_5$ are accepting states. Our goal is to derive closed-form REs for $q_3$ and $q_5$ in terms of the input alphabet $\{a,b\}$ only.

\textbf{Step 2: Solve for $q_0$ using Arden's theorem.} Equation~\eqref{eq:q0} has the form $R = Q + RP$ with $R = q_0$, $Q = \epsilon$, and $P = b$. Since $b$ does not contain $\epsilon$, Arden's theorem yields: $q_0 = \epsilon \cdot b^* = b^*$

\textbf{Step 3: Solve for $q_1$.} Substitute $q_2 = q_1 b$ from Equation~\eqref{eq:q2} into Equation~\eqref{eq:q1}:
\[
\begin{aligned}
q_1 &= q_0 a + q_1 a + q_2 a = q_0 a + q_1 a + (q_1 b)a \\
    &= q_0 a + q_1(a + ba)
\end{aligned}
\]
Applying Arden's theorem with $R = q_1$, $Q = q_0 a$, $P = (a+ba)$: $q_1 = q_0 a(a+ba)^* = b^* a(a+ba)^*$. 

\textit{Following a similar application of Arden's theorem, we get the following REs for the final states:} $q_3 = b^* a (a+ba)^* b b (a+b)^* $ and $q_5 = b^* b b^* a (a+b)^*$

\textbf{Final language:} Therefore, the language recognized by the NFA is: $L_3 = q_3 \cup q_5$. Hence, 
\[
L_3 = b^*a(a+ba)^*bb(a+b)^* + b^*bb^*a(a+b)^*
\]
The resulting expression exhibits deep nesting (concatenations of unions containing Kleene stars) and non-standard structure unlikely to match textbook examples. Figure~\ref{fig:L3}(b) shows the minimal DFA for $L_3$, obtained by converting the NFA via subset construction and state minimization. Using this procedure, we generated $120$ problems spanning a range of structural complexities. Each random NFA seed produces a distinct RE, ensuring comprehensive coverage of expression patterns while maintaining verifiable correctness through algorithmic DFA construction\footnote{All datasets, file formats, and additional implementation details are provided in Appendix \ref{app:datasets}.}.

\subsection{Dataset Statistics and Difficulty level}

We categorize all $180$ unseen problems by difficulty based on structural complexity: (i) \textbf{Easy ($41$ problems, $22.8$\%):} Simple expressions with shallow nesting (e.g., $L_4 = b(a+b)^*ab$ -- $1$ Kleene star, $1$ union, $3$ concatenations, minimal DFA has $\leq 5$ states). (ii) \textbf{Medium ($65$ problems, $36.1$\%):} Moderate compositionality with $2-4$ nested operators (e.g., $L_5 = (a^*b + ((a^*a + a^*b)b^*)a + (((a^*a + a^*b)b^*a)b^*)b)a^*$; minimal DFA has $6-12$ states). (iii) \textbf{Hard ($74$ problems, $41.1$\%):} Deep nesting requiring global consistency tracking (e.g., $L_6 = (((a^*b)(a + b((\epsilon + a)(ba)^*(a+bb) + b)b^*a)^*)b((\epsilon+a)(ba)^*(a+bb)+b))b^*$; minimal DFA has $\geq 13$ states). For manually composed problems, difficulty reflects the number of interacting constraints ($1 - 2$ for easy, $3 - 4$ for medium, $5 +$ for hard). For Arden-generated problems, difficulty corresponds to RE nesting depth and the number of states in the DFA. Table~\ref{tab:dataset_stats} summarizes the complete benchmark. The difficulty distribution ensures comprehensive evaluation across complexity levels, with a slight bias toward hard problems (41\%) to stress-test reasoning capabilities, for detailed classification see Appendix~\ref{app:EasyMed}.

\begin{table}[t]
\centering
\scriptsize
\begin{tabular}{lcccc}
\toprule
\textbf{Dataset Component} & \textbf{Count} & \textbf{Easy} & \textbf{Med} & \textbf{Hard} \\
\midrule
\textbf{Unseen construction} & \textbf{180} & \textbf{41} & \textbf{65} & \textbf{74} \\
\quad Manual composition & 60 & 15 & 22 & 23 \\
\quad Arden generation & 120 & 26 & 43 & 51 \\
\bottomrule
\end{tabular}
\caption{\small{Dataset statistics and difficulty categorization.}}
\label{tab:dataset_stats}
\end{table}

\noindent\textbf{Validation of Novelty.} To ensure problems are truly unseen, we: (1) manually searched for exact and structurally similar matches in top Google results and online repositories, (2) verified that Arden-generated regular expressions do not match patterns in standard textbooks, and (3) confirmed that manual problems combine constraints in non-standard ways. While we cannot guarantee complete absence from all pretraining corpora, the controlled generation process and verification steps provide strong evidence of novelty.

\section{Evaluation Methodology}
\label{sec:methods}
Here, all models are accessed via official APIs with temperature set to $0$ for deterministic decoding, and all experiments use identical prompt templates to ensure fair comparison \footnote{Additional details regarding determinism controls, decoding settings, and the runtime environment are provided in Appendices \ref{app:determinism} and \ref{app:runtime}.}. We evaluate five prompting configurations spanning zero-shot, explicit reasoning, multi-branch exploration, and guided self-correction:

\noindent\textbf{Direct Input-Output (Zero-Shot).} Models receive only the problem specification (RE or natural language) and must output a complete DFA with no intermediate reasoning or worked examples~\cite{kojima2022}. This baseline isolates pure construction capability without scaffolding.

\noindent\textbf{Chain-of-Thought (CoT).} Models are instructed to reason step-by-step before producing the final DFA~\cite{wang2024c}. For DFA construction, this naturally decomposes into: (1) interpreting the language specification, (2) identifying required states and their semantic roles, and (3) designing transitions ensuring correct acceptance/rejection. For example, given $L_7 = (a+b)^*a(a+b)^*$, CoT prompting encourages models to explicitly reason: ``State $q_0$ tracks strings without `a'; state $q_1$ (accepting) tracks strings with at least one `a'; transition $\delta(q_0, a) = q_1$ handles first `a'; etc.''.

\noindent\textbf{Chain-of-Thought (One-Shot).} Extends CoT by providing a single worked example (problem + solution + reasoning trace) before the target problem. This tests whether explicit demonstration reduces ambiguity in reasoning structure.

\noindent\textbf{Tree-of-Thought (ToT).} Decomposes DFA construction into four distinct reasoning branches corresponding to standard automata-theoretic methods~\cite{10.5555/3666122.3666639}. Models are prompted to explore multiple construction approaches in parallel: (i) \textbf{Direct (Intuitive):} Interpret the regular expression semantically, identify necessary states based on language properties, and design transitions directly. Tests high-level symbolic reasoning without algorithmic scaffolding. (ii) \textbf{Minimization-Based:} Construct an initial DFA (possibly with redundant states), then merge equivalent states via partition refinement. Tests whether models understand state equivalence beyond local transition correctness. (iii) \textbf{Derivative-Based:} Apply Brzozowski's method~\cite{sipser13}, where each state corresponds to a distinct derivative of the regular expression with respect to input prefixes. Tests symbolic algebraic manipulation and normalization of equivalent expressions. (iv) \textbf{Thompson's Construction:} Follow the algorithmic pipeline: convert regular expression to $\epsilon$-NFA via structural recursion, then determinize via subset construction. Tests procedural correctness on multi-stage formal algorithms. Lexical analyser tools Lex, Flex follow this approach to automatically create DFA from RE with limitations \cite{10.5555/1177220}. These branches span fundamentally different reasoning styles (semantic interpretation, equivalence optimization, algebraic manipulation, algorithmic execution), enabling fine-grained diagnosis of where models succeed or fail. All prompt templates are provided in Appendix~\ref{app:prompts}.

\noindent\textbf{Hint-Based Self-Correction Protocol.} For problems answered incorrectly under direct prompting, we evaluate whether models can self-correct when provided structured feedback. We introduce a three-stage hint protocol with increasing levels of guidance: \textbf{Stage 1: Counterexample Feedback.} The model receives concrete counterexamples exposing errors in its DFA: \textbf{False negatives:} Strings in $L$ that the DFA rejects;  \textbf{False positives:} Strings not in $L$ that the DFA accepts. Counterexamples are selected to cover all major structural paths through the automaton, ensuring errors cannot be fixed by local patches. \textbf{Stage 2: Error Localization.} If errors persist, the model is informed \emph{which reasoning stage} is incorrect (language interpretation, state design, or transition) along with a high-level description (e.g.,~`Your DFA incorrectly simplifies the Kleene star constraint'). \textbf{Stage 3: Explicit Error Disclosure.} If the model still fails, it receives the exact error location and nature (e.g.,~`Missing transition: $\delta(q_2, a) \rightarrow q_4$ to handle aba'). This progressive disclosure tests whether failures stem from shallow misinterpretation (correctable with counterexamples), incomplete reasoning (correctable with localization), or fundamental inability to maintain symbolic consistency (uncorrectable even with explicit guidance), Appendix~\ref{app:prompts} contains hint templates.

\noindent\textbf{Validation and Grading.} All DFA outputs are validated through a two-stage pipeline combining automated behavioural testing and manual structural inspection: (i) \textbf{Automated Validation:} A custom validator checks: (a) \textit{syntactic correctness}: valid JSON schema, well-formed state/transition structure; (b) \textit{totality}: every state has exactly one outgoing transition per input symbol; and (c) \textit{behavioural equivalence}: the DFA accepts the same language as the ground-truth RE. Behavioural equivalence is tested via exhaustive enumeration of all strings up to length $6$ and random sampling of $2000$ strings of length $7-15$. Any mismatch triggers explicit counterexample logging. (ii) \textbf{Manual Inspection:} For all DFAs, at least two independent reviewers from the research team manually verify: (a) correct interpretation of the language specification, (b) meaningful state semantics aligned with the RE, and (c) valid transition logic. All grading is conducted in a model blinded manner to prevent bias. A DFA is marked correct if and only if it passes both tests (see Appendices~\ref{app:validation} and~\ref{app:execution_code} for validation codes, evaluation scripts and execution scripts).

\section{Results and Analysis}

\noindent\textbf{Results.} Table~\ref{Table1} reports DFA construction success rates across datasets, prompting strategies, and models. (i) \textbf{Seen vs.\ Unseen:} In contrast with the results of seen dataset (Section~\ref{sec:seen_results}), performance drops sharply on the \emph{Unseen DFA Construction} dataset across all models. Under direct prompting, Grok-$4.1$-fast-reasoning performs best, followed by Gemini-$2.5$-Flash and GPT-$5.1$. This corresponds to a degradation of $30$--$64$ \% relative to seen tasks. Crucially, the task formulation is identical across seen and unseen settings. The observed performance gap therefore isolates a failure of \emph{compositional generalization} rather than task understanding. (ii) \textbf{CoT Effects:} CoT consistently degrades performance on unseen DFA construction across all models. Here, explicit step-by-step reasoning increases effective state-space complexity and amplifies the impact of early semantic errors. To assess whether examples mitigate this failure mode, we evaluate \emph{CoT (One-Shot)} prompting. Providing a single worked example improves performance relative to standard CoT. (iii) \textbf{ToT Effects:} ToT yields the strongest performance on complex unseen instances. Gemini-$2.5$, GPT-$5.1$ improve by $24.67$,~$3.43$ \% over direct prompting and $30.67$,~$7.43$ \% over CoT prompting respectively. In contrast, Grok-$4.1$-fast-reasoning frequently fails to return an output within the fixed inference-time budget \footnote{A detailed analysis of timeout behaviour, inference cost, and deployability trade-offs is provided in Section~\ref{app:limitations}.}. 

\begin{table}[t]
\centering
\scriptsize
\setlength{\tabcolsep}{3pt}
\begin{tabular}{l l c c c}
\toprule
\textbf{Dataset} & \textbf{Prompting} & \textbf{GPT-5.1} & \textbf{Grok-4.1} & \textbf{Gemini-2.5} \\
\midrule
Seen DFA   & Zero-Shot     & 84.2\%  & 89.5\%  & 85.0\% \\
\midrule
Unseen DFA & Zero-Shot     & 20.67\% & 59.12\% & 29.33\% \\
           & CoT              & 16.67\% & 51.90\% & 23.33\% \\
           & CoT (One-Shot)   & 19.33\% & 55.87\% & 28.10\% \\
           & ToT              & 24.10\% & --      & 54.00\% \\
\bottomrule
\end{tabular}
\caption{\small{DFA construction success rates across datasets, prompting strategies, and models. ``--'' denotes inference-time timeouts.}}
\label{Table1}
\end{table}

\noindent\textbf{Analysis.} Across models and prompting strategies, the dominant failure modes arise from difficulties in maintaining \emph{globally consistent symbolic structure}, rather than from a lack of formal knowledge. To illustrate, we analyze following representative outputs under the four ToT framework methods (which subsumes Direct and CoT). (i) \textbf{Thompson Construction:} Here, LLMs generally succeed on moderately complex REs with limited nesting because this method is fully algorithmic. However, for complex expressions involving deep nesting, the intermediate $\varepsilon$-NFA grows rapidly in size. This \emph{state explosion} acts as a reason behind the failure of LLMs. Importantly, these failures do not reflect a misunderstanding of method, but rather a limitation in reliably managing large, densely connected intermediate structures. (ii) \textbf{Direct Construction:} Here, LLMs attempt to construct DFAs by informally reasoning about the language constraints. This setting exhibits following recurring failure modes. \emph{(a) Partial or oversimplified language interpretation.} To produce a correct DFA, the first most logically important step demands correct understanding about the language. However, there are many evidences where LLMs lack in this step (e.g.,~for $L_{8}$ = $a^{*}(a+b)(a+b)^{*}a$, LLM develops wrong understanding `ending with a', ignoring presence of $(a+b)$). The same is also applicable for `Kleene star'. \emph{(b) Inconsistency between a model's stated understanding and its final construction.} (e.g.,~starts with correct understanding `strings ending with $ab$', but returns answer for `ending with $ba$'). It indicates difficulty in maintaining symbolic invariants across construction. \emph{(c) Failure to preserve constraints under concatenation.} (e.g.,~fails to handle $L_8$, but capable to handle unit components, namely $a^{*}$, $(a+b)$, $(a+b)^{*}$, and $a$). \emph{(d) Over-acceptance of strings outside the target language.} Models often validate their constructions using a finite set of accepted strings while ignoring rejection cases. In addition, LLMs often introduce unreachable or redundant states, which further invalidate the answer \footnote{Appendix~\ref{app:failure-modes} contains a gallery of representative errors.}. (iii) \textbf{Minimization-Based Construction.} Here, most errors originate before minimization is applied and propagates during the algorithmic minimization approach. (iv) \textbf{Derivative-Based Construction.} LLMs frequently fail to normalize semantically equivalent RE derivatives, leading to the creation of spurious DFA states. (e.g.,~for $L_9 = b(a+b)^*ab$\footnote{see Appendix~\ref{app:derivative_L8} for model-generated output.} the derivative $(a+b)^*ab + \varepsilon$ is a correct intermediate step reflecting nullability. However, LLMs frequently replace $\varepsilon$ with $(a+b)^*$, yielding the incorrect intermediate RE $(a+b)^*ab + (a+b)^*$. This RE normalizes to $(a+b)^*$, thus discards the original language constraints). Importantly, this failure shows an inability to perform basic algebraic simplification and language containment reasoning.

\noindent\textbf{Hint-Based Protocol Analysis.} The hint-based protocol is designed to evaluate whether LLMs can recover from incorrect DFA constructions when provided with structured guidance. Table~\ref{Table2} summarizes model performance across difficulty levels and hint stages. (i) \textbf{Easy:} Most errors are resolved after the \emph{first hint} across all models, indicating that failures at this level primarily arise from superficial misinterpretations rather than fundamental reasoning errors. (ii) \textbf{Medium:} Here, recovery is distributed across multiple hint stages rather than concentrated at a single level of guidance. Moreover, a non-trivial fraction remain incorrect even after the final hint. (iii) \textbf{Difficult:} A large proportion of problems remain unsolved even after all hints are provided, particularly for GPT-$5.1$ and Gemini-$2.5$-Flash. The persistence of unsolved cases suggests that, for complex REs, guidance is insufficient to overcome underlying limitations in symbolic state-transition tracking, and global automaton structure. Overall, the hint-based analysis demonstrates that while all models can leverage guidance to correct shallow errors, their ability to recover degrades rapidly as DFA complexity increases. Hints primarily assist in resolving local misunderstandings, but they do not reliably enable correction of globally inconsistent or structurally flawed automata.

\begin{table}[t]
\centering
\scriptsize
\setlength{\tabcolsep}{3pt}
\resizebox{\columnwidth}{!}{%
\begin{tabular}{l|ccc|ccc|ccc}
\toprule
\textbf{Stage}
& \multicolumn{3}{c|}{\textbf{GPT-5.1}}
& \multicolumn{3}{c|}{\textbf{Grok-4.1}}
& \multicolumn{3}{c}{\textbf{Gemini-2.5}} \\

\cmidrule(lr){2-4} \cmidrule(lr){5-7} \cmidrule(lr){8-10}
& Easy & Medium & Difficult
& Easy & Medium & Difficult
& Easy & Medium & Difficult \\
\midrule

First Hint
& 60.13\% & 10.13\% & 30.76\%
& 85.25\% & 39.20\% & 52.93\%
& 57.45\% & 49.13\% & 31.79\% \\

Second Hint
& 15.50\% & 16.10\% & 15.38\%
& 14.75\% & 27.67\% & 7.83\%
& 42.55\% & 12.47\% & 6.88\% \\

Final Hint
& 24.37\% & 36.17\% & --
& -- & 13.33\% & --
& -- & 12.37\% & 16.31\% \\

Not Solved
& -- & 37.60\% & 53.86\%
& -- & 19.80\% & 39.24\%
& -- & 26.03\% & 45.02\% \\
\bottomrule
\end{tabular}}
\caption{\small Hint-based DFA construction performance.}
\label{Table2}
\end{table}

\section{Concluding Remarks}

This work presents a systematic evaluation of LLMs on DFA construction tasks, with an emphasis on symbolic correctness rather than surface-level pattern matching. We find that while current LLMs reliably reproduce familiar constructions, their performance degrades substantially on compositionally complex inputs. Our analysis shows that these failures are systematic rather than random. These limitations persist across prompting strategies indicating structural weaknesses in symbolic reasoning rather than prompt-specific shortcomings. The hint-based framework demonstrates that limited self-correction is possible for simpler errors; however, deeper semantic inconsistencies are rarely resolved. By introducing a principled benchmark with controlled novelty and by identifying recurring failure modes, this work provides a foundation for future research on improving symbolic robustness through improved evaluation methodologies, targeted training, or hybrid symbolic--neural approaches.

\section{Limitations}
\label{app:limitations}

This study has several limitations that should be considered when interpreting
the results.

\paragraph{Model access and evaluation constraints.} All experiments were conducted via public APIs or standard model interfaces with
fixed inference-time budgets. As a result, some models---most notably
Grok-4.1-fast-reasoning under Tree-of-Thought prompting---frequently failed to
return outputs within the allotted budget. Such cases were treated as
unsuccessful attempts, reflecting practical interface and deployability
constraints rather than definitive limitations of the underlying reasoning
capabilities.

\paragraph{Token and state-space scalability.} DFA construction from complex regular expressions often induces large intermediate representations, particularly under Chain-of-Thought and Tree-of-Thought prompting. Token budget exhaustion and implicit state-space explosion limit the reliability of these methods for highly nested or compositional expressions, even when the underlying construction is theoretically well defined.

\paragraph{Finite validation horizon.} Behavioral equivalence between generated DFAs and ground-truth regular expressions is tested using exhaustive enumeration only up to a bounded string length, supplemented by randomized testing for longer strings. While this approach provides strong empirical assurance, it does not constitute a formal proof of equivalence for all possible strings. As a result, rare counterexamples beyond the tested length range may remain undetected, potentially leading to overestimation of construction correctness.

\paragraph{Dataset format constraints.} All DFA construction datasets are released in PDF format, containing problem statements, reference transition tables, and DFA diagrams. This design choice prioritizes human interpretability and preserves pedagogical structure, but it requires manual or semi-automated parsing for downstream reuse and limits direct machine-readability.

\paragraph{Scope of tasks and formalisms.} The experiments focus exclusively on regular expressions, DFA/NFA construction, and Arden’s theorem--based inverse synthesis. Consequently, the reported results may not generalize to richer automata models, non-regular languages, or other formal systems. Accordingly, our findings should not be generalized to richer formalisms such as context-free grammars, pushdown automata, or Turing-complete models.

\paragraph{Model variability across interfaces.} Identical prompts issued via APIs and web interfaces can yield different outputs due to undocumented system-level differences. All reported results correspond to API-based evaluations and should not be interpreted as exact replicas of web interface behavior.

\paragraph{Absence of fine-tuning or adaptive prompting.} No model-specific fine-tuning, adaptive decoding, or prompt optimization was applied beyond the fixed prompting protocols described in the paper. While this ensures controlled comparability across models, it may underestimate achievable performance under specialized or tuned evaluation settings.

These limitations primarily affect the external validity of our conclusions, but do not alter the observed relative performance trends across models and prompting strategies.

\section{Ethical Considerations}
\label{app:ethics}

This work evaluates the capability of large language models (LLMs) to perform
formal language--theoretic reasoning tasks, specifically deterministic finite
automaton (DFA) construction and related inverse problems. The study does not
involve human subjects, personal data, or sensitive information.

\paragraph{Dataset sourcing.} The datasets used in this study consist exclusively of symbolic, mathematical objects, including regular expressions, DFAs, NFAs, transition tables, and multiple-choice or true/false questions.

The Knowledge Checking dataset and the Seen DFA Construction dataset were sourced
from publicly available educational materials and standard automata theory
problem sets accessible online. These materials are commonly used for teaching
and assessment and do not carry restrictive licenses.

The Unseen DFA Construction dataset, including the Mathematical Engineering and
Mathematical Art subsets, was manually constructed by the authors and was not
sourced from online repositories. All datasets are released with an explicit
data license in the accompanying repository.

All newly created datasets are released under the Creative Commons Attribution 4.0 International (CC BY 4.0) license. Publicly sourced educational datasets are provided for research and evaluation purposes only, with original copyrights retained by their respective owners. Full license text and terms of use are provided in the accompanying repository.

\paragraph{Model usage and evaluation.} All evaluated models were accessed exclusively through their official public APIs under standard usage conditions. No attempts were made to extract proprietary model internals, training data, or hidden reasoning traces. Prompts were designed solely to elicit task performance and did not request personal, sensitive, or copyrighted information.

\paragraph{Reproducibility and transparency.} To promote transparency and responsible research practice, we release the full set of prompts, experimental scripts, validation code, and representative outputs. Automated validation and targeted human inspection were used to reduce the risk of reporting spurious or misleading results.

\paragraph{Potential risks and misuse.} The tasks studied are purely formal and mathematical in nature, and the released artifacts are not expected to enable harmful applications or deployment-facing systems. A plausible risk lies in misinterpretation or overgeneralization of the results, for example if observed failures or successes are taken as evidence of general reasoning ability or deficiency across broader linguistic or real-world domains. We explicitly limit our claims to regular-language formalisms and symbolic automata construction and caution against extrapolation beyond these settings.

\paragraph{Use of AI assistants.} AI assistants were used in a limited capacity during the development and debugging of auxiliary code. They were not used to generate datasets, ground-truth solutions, experimental results, analyses, or conclusions, and did not contribute to the scientific claims of this work.

Overall, we identify no significant ethical risks associated with this study and
adhere to conference guidelines on data sourcing, transparency, and responsible
reporting.

\bibliography{custom1}

\appendix

\section{Appendix : Model-Specific DFA Outputs}
\label{app:dfa_outputs}

This appendix reports the exact DFA constructions produced by each evaluated
model for the seen task ($L_1$) and the unseen task ($L_2$). All outputs are
shown exactly as returned by the models, without post-processing or correction.
\subsection{Seen Problem ($L_1$)}

\subsubsection{GPT-5.1 Output for $L_1$}
\begin{lstlisting}[language=json]
  {
    "states": [
      "q0",
      "q1",
      "q2",
      "q3",
      "q4",
      "q5",
      "q6",
      "q7"
    ],
    "alphabet": [
      "a",
      "b"
    ],
    "start_state": "q0",
    "accept_states": [
      "q4",
      "q5",
      "q6",
      "q7"
    ],
    "transitions": {
      "q0": {
        "a": "q1",
        "b": "q0"
      },
      "q1": {
        "a": "q3",
        "b": "q2"
      },
      "q2": {
        "a": "q5",
        "b": "q4"
      },
      "q3": {
        "a": "q7",
        "b": "q6"
      },
      "q4": {
        "a": "q1",
        "b": "q0"
      },
      "q5": {
        "a": "q3",
        "b": "q2"
      },
      "q6": {
        "a": "q5",
        "b": "q4"
      },
      "q7": {
        "a": "q7",
        "b": "q6"
      }
    }
  }
\end{lstlisting}

\subsubsection{Grok-4-1-fast-reasoning Output for $L_1$}
\begin{lstlisting}[language=json]
{
  "states": [
    "q0",
    "q1",
    "q2",
    "q3",
    "q4",
    "q5",
    "q6",
    "q7"
  ],
  "alphabet": [
    "a",
    "b"
  ],
  "start_state": "q0",
  "accept_states": [
    "q4",
    "q5",
    "q6",
    "q7"
  ],
  "transitions": {
    "q0": {
      "a": "q1",
      "b": "q0"
    },
    "q1": {
      "a": "q3",
      "b": "q2"
    },
    "q2": {
      "a": "q5",
      "b": "q4"
    },
    "q3": {
      "a": "q7",
      "b": "q6"
    },
    "q4": {
      "a": "q1",
      "b": "q0"
    },
    "q5": {
      "a": "q3",
      "b": "q2"
    },
    "q6": {
      "a": "q5",
      "b": "q4"
    },
    "q7": {
      "a": "q7",
      "b": "q6"
    }
  }
}
\end{lstlisting}

\subsubsection{Gemini-2.5-flash Output for $L_1$}
\begin{lstlisting}[language=json]
{
  "states": [
    "q0",
    "q1",
    "q2",
    "q3",
    "q4",
    "q5",
    "q6",
    "q7"
  ],
  "alphabet": [
    "a",
    "b"
  ],
  "start_state": "q0",
  "accept_states": [
    "q4",
    "q5",
    "q6",
    "q7"
  ],
  "transitions": {
    "q0": {
      "a": "q1",
      "b": "q0"
    },
    "q1": {
      "a": "q2",
      "b": "q3"
    },
    "q2": {
      "a": "q4",
      "b": "q5"
    },
    "q3": {
      "a": "q6",
      "b": "q7"
    },
    "q4": {
      "a": "q4",
      "b": "q5"
    },
    "q5": {
      "a": "q6",
      "b": "q7"
    },
    "q6": {
      "a": "q2",
      "b": "q3"
    },
    "q7": {
      "a": "q1",
      "b": "q0"
    }
  }
}
\end{lstlisting}

\subsection{Unseen Problem ($L_2$)}

\subsubsection{GPT-5.1 Output for $L_2$}
\begin{lstlisting}[language=json]
{
  "states": [
    "q0",
    "q1",
    "q2",
    "q3",
    "q4",
    "q5",
    "q6",
    "q7",
    "q8",
    "q9",
    "q10",
    "q11",
    "q12",
    "q13",
    "q14",
    "q15",
    "q16",
    "q17",
    "q18",
    "q19",
    "q20",
    "q21",
    "q22",
    "q23",
    "q24",
    "q25",
    "q26",
    "q27",
    "q28",
    "q29",
    "q30",
    "q31",
    "qDead"
  ],
  "alphabet": [
    "a",
    "b"
  ],
  "start_state": "q0",
  "accept_states": [
    "q24",
    "q25",
    "q26",
    "q27",
    "q28",
    "q29",
    "q30",
    "q31"
  ],
  "transitions": {
    "q0": {
      "a": "q1",
      "b": "q2"
    },
    "q1": {
      "a": "q3",
      "b": "q4"
    },
    "q2": {
      "a": "q1",
      "b": "qDead"
    },
    "q3": {
      "a": "q5",
      "b": "q6"
    },
    "q4": {
      "a": "q3",
      "b": "q7"
    },
    "q5": {
      "a": "q8",
      "b": "q9"
    },
    "q6": {
      "a": "q5",
      "b": "q10"
    },
    "q7": {
      "a": "q6",
      "b": "q11"
    },
    "q8": {
      "a": "q12",
      "b": "q13"
    },
    "q9": {
      "a": "q8",
      "b": "q14"
    },
    "q10": {
      "a": "q9",
      "b": "q15"
    },
    "q11": {
      "a": "q10",
      "b": "q16"
    },
    "q12": {
      "a": "q17",
      "b": "q18"
    },
    "q13": {
      "a": "q12",
      "b": "q19"
    },
    "q14": {
      "a": "q13",
      "b": "q20"
    },
    "q15": {
      "a": "q14",
      "b": "q21"
    },
    "q16": {
      "a": "q15",
      "b": "q22"
    },
    "q17": {
      "a": "q23",
      "b": "q24"
    },
    "q18": {
      "a": "q17",
      "b": "q25"
    },
    "q19": {
      "a": "q18",
      "b": "q26"
    },
    "q20": {
      "a": "q19",
      "b": "q27"
    },
    "q21": {
      "a": "q20",
      "b": "q28"
    },
    "q22": {
      "a": "q21",
      "b": "q29"
    },
    "q23": {
      "a": "q23",
      "b": "q30"
    },
    "q24": {
      "a": "q23",
      "b": "q31"
    },
    "q25": {
      "a": "q24",
      "b": "q31"
    },
    "q26": {
      "a": "q25",
      "b": "q31"
    },
    "q27": {
      "a": "q26",
      "b": "q31"
    },
    "q28": {
      "a": "q27",
      "b": "q31"
    },
    "q29": {
      "a": "q28",
      "b": "q31"
    },
    "q30": {
      "a": "q30",
      "b": "q31"
    },
    "q31": {
      "a": "q30",
      "b": "q31"
    },
    "qDead": {
      "a": "qDead",
      "b": "qDead"
    }
  }
}
\end{lstlisting}

\subsubsection{Grok-4-1-fast-reasoning Output for $L_2$}
\begin{lstlisting}[language=json]
{

  "states": [
    "q0",
    "q1",
    "q2",
    "q3",
    "q4",
    "q5",
    "q6",
    "q7",
    "q8",
    "q9",
    "q10",
    "q11",
    "q12",
    "q13",
    "q14",
    "q15",
    "q16",
    "q17",
    "q18",
    "q19",
    "q20",
    "q21",
    "q22",
    "q23",
    "q24",
    "q25",
    "q26",
    "q27",
    "q28"
  ],
  "alphabet": [
    "a",
    "b"
  ],
  "start_state": "q0",
  "accept_states": [
    "q13",
    "q14",
    "q15",
    "q16",
    "q17",
    "q18",
    "q19",
    "q20"
  ],
  "transitions": {
    "q0": {
      "a": "q1",
      "b": "q2"
    },
    "q1": {
      "a": "q3",
      "b": "q4"
    },
    "q2": {
      "a": "q5",
      "b": "q12"
    },
    "q3": {
      "a": "q6",
      "b": "q7"
    },
    "q4": {
      "a": "q8",
      "b": "q9"
    },
    "q5": {
      "a": "q10",
      "b": "q11"
    },
    "q6": {
      "a": "q13",
      "b": "q14"
    },
    "q7": {
      "a": "q15",
      "b": "q16"
    },
    "q8": {
      "a": "q17",
      "b": "q18"
    },
    "q9": {
      "a": "q19",
      "b": "q20"
    },
    "q10": {
      "a": "q21",
      "b": "q22"
    },
    "q11": {
      "a": "q23",
      "b": "q24"
    },
    "q12": {
      "a": "q12",
      "b": "q12"
    },
    "q13": {
      "a": "q13",
      "b": "q14"
    },
    "q14": {
      "a": "q15",
      "b": "q16"
    },
    "q15": {
      "a": "q17",
      "b": "q18"
    },
    "q16": {
      "a": "q19",
      "b": "q20"
    },
    "q17": {
      "a": "q21",
      "b": "q22"
    },
    "q18": {
      "a": "q23",
      "b": "q24"
    },
    "q19": {
      "a": "q25",
      "b": "q26"
    },
    "q20": {
      "a": "q27",
      "b": "q28"
    },
    "q21": {
      "a": "q13",
      "b": "q14"
    },
    "q22": {
      "a": "q15",
      "b": "q16"
    },
    "q23": {
      "a": "q17",
      "b": "q18"
    },
    "q24": {
      "a": "q19",
      "b": "q20"
    },
    "q25": {
      "a": "q21",
      "b": "q22"
    },
    "q26": {
      "a": "q23",
      "b": "q24"
    },
    "q27": {
      "a": "q25",
      "b": "q26"
    },
    "q28": {
      "a": "q27",
      "b": "q28"
    }
  }
}

\end{lstlisting}

\subsubsection{Gemini-2.5-flash Output for $L_2$}
\begin{lstlisting}[language=json]
{
  "states": [
    "q0",
    "q1",
    "q2",
    "q3",
    "q4",
    "q5",
    "q6",
    "q7",
    "q8",
    "q9",
    "q10",
    "q11",
    "q12",
    "q13",
    "q14",
    "q15"
  ],
  "alphabet": [
    "a",
    "b"
  ],
  "start_state": "q0",
  "accept_states": [
    "q7",
    "q8",
    "q9",
    "q10"
  ],
  "transitions": {
    "q0": {
      "a": "q2",
      "b": "q1"
    },
    "q1": {
      "a": "q5",
      "b": "q15"
    },
    "q2": {
      "a": "q3",
      "b": "q4"
    },
    "q3": {
      "a": "q7",
      "b": "q8"
    },
    "q4": {
      "a": "q9",
      "b": "q10"
    },
    "q5": {
      "a": "q11",
      "b": "q12"
    },
    "q6": {
      "a": "q13",
      "b": "q14"
    },
    "q7": {
      "a": "q7",
      "b": "q8"
    },
    "q8": {
      "a": "q9",
      "b": "q10"
    },
    "q9": {
      "a": "q11",
      "b": "q12"
    },
    "q10": {
      "a": "q13",
      "b": "q14"
    },
    "q11": {
      "a": "q7",
      "b": "q8"
    },
    "q12": {
      "a": "q9",
      "b": "q10"
    },
    "q13": {
      "a": "q11",
      "b": "q12"
    },
    "q14": {
      "a": "q13",
      "b": "q14"
    },
    "q15": {
      "a": "q15",
      "b": "q15"
    }
  }
}
\end{lstlisting}

\section{Appendix : Datasets}
\label{app:datasets}

This appendix documents the datasets used in our experiments, including their
composition, sizes, formats, and public availability. Detailed dataset
construction procedures, design motivations, and validation protocols are
described in the main paper.

\subsection{Overview}

We evaluate large language model (LLM) performance using four datasets spanning
factual knowledge, seen DFA construction, and two categories of unseen DFA
construction tasks. All datasets are grounded in formal language theory and
deterministic finite automata. Ground-truth answers are verified through
manual construction and automated validation procedures described in the main
paper.

\subsection{Dataset Composition and Sizes}

\begin{itemize}
    \item \textbf{Knowledge Checking Dataset} (50 questions): \\
    This dataset consists of multiple-choice and true/false questions assessing
    factual knowledge of core Theory of Computation concepts, including regular
    languages, deterministic and nondeterministic finite automata, and DFA
    minimization. Each question admits a single unambiguous correct answer.

    \item \textbf{Seen DFA Construction Dataset} (90 questions): \\
    This dataset contains regular expression–to–DFA and language–to–DFA
    construction problems whose structural patterns closely resemble commonly
    available textbook exercises and publicly accessible online examples.
    For each problem, we additionally provide explicit evidence demonstrating
    the availability of structurally identical or equivalent questions on the
    public internet, serving as proof of prior exposure.

   \item \textbf{Unseen DFA Construction Dataset} (180 questions): \\
This dataset is designed to evaluate generalization beyond memorized patterns and
is intentionally constructed to avoid overlap with common instructional
examples. It is divided into two subsets that differ in their generative
principles:

\begin{itemize}

\item \textbf{Mathematical Art Subset} (60 questions):\\
This subset consists of problems generated by combining multiple interacting
constraints, including structural restrictions on automata, conditional
dependencies between symbols, and narrative-based task formulations. The
resulting languages are intentionally highly irregular and non-canonical,
designed to stress symbolic consistency, constraint integration, and
long-horizon reasoning.

To design the languages in this unseen subset, we employ the following
principled construction strategies:

\begin{enumerate}

\item[(a)] \textbf{Use of product construction.}
Multiple constraints are combined using product-style constructions (e.g.,
repeated use of conjunctions and disjunctions to increase constraint
complexity); for example,

$L_{a} =$ {\tt
\{ Construct a DFA that accepts the set of all strings over \{a,b\}
where it starts with either `aba' or `bab' and ends with `bab' or `aba'
respectively. \}
}

$L_{b} =$ {\tt
\{ Construct a DFA that accepts the set of all strings over \{a,b\}
where `ab' should be followed by `ba' and `aa' should be followed by `bb'
and the total count of `a's and `b's is even. \}
}

\item[(b)] \textbf{Multiple interacting constraints.}
Independent constraints (e.g., prefix conditions paired with counting or
exclusion constraints) are imposed simultaneously, preventing purely local
reasoning from yielding a correct construction; for example,

$L_{c} =$ {\tt
\{ Construct a DFA over \{a,b\} that accepts all strings in which
the fourth-last symbol from the end must be `a' and the substring `bb'
does not appear before any `a'. \}
}

$L_{d} =$ {\tt
\{ Construct a DFA over \{a,b,c\} such that there exists a block of
A's of length no more than 5, immediately followed by a block of B's that
is twice the length of the A block; the character C may appear anywhere
before or after these two consecutive blocks with $|C| \equiv 0 \pmod{3}$,
and no C appears between the two AB blocks. \}
}

\item[(c)] \textbf{Structural restrictions.}
Structural constraints such as adjacency rules, forbidden substrings, or
positional patterns are enforced, requiring precise tracking of symbol
relationships; for example,

$L_{e} =$ {\tt
\{ Construct a DFA over \{0,1\} that accepts a string w such that the digit
obtained by the XOR operation of the first three digits is equal to the
digit present at the end of the string w. \}
}

$L_{f} =$ {\tt
\{ Construct a DFA over \{0,1\} such that the string length is greater than
or equal to 4, and in each substring of length 4 the first two bits are
interpreted as a decimal number and the last two bits are interpreted as
another decimal number, and their multiplication is divisible by 4. \}
}

\item[(d)] \textbf{Narrative-based tasks.}
Narrative-driven problems are formulated using real-world or game-based
scenarios, which must be translated into precise symbolic automaton
constraints; for example,

$L_{g} =$ {\tt
\{ In chess, consider the Ruy-Lopez opening. States correspond to positions
of pieces on the board. The input alphabet is defined as
Sigma = \{pxyi\}, where p denotes the player (white or black), x denotes the
piece, y the square alphabet, and i the square number. A move such as a
black knight moving to c6 is written as bkc6. Assuming standard chess rules
with alternating turns and a fixed initial board configuration, construct a
minimized DFA whose accepting paths correspond to the Ruy-Lopez opening
achieved in exactly five moves, excluding the two initial moves. \}
}

\begin{figure}[t]
\centering
\includegraphics[width=0.8\columnwidth]{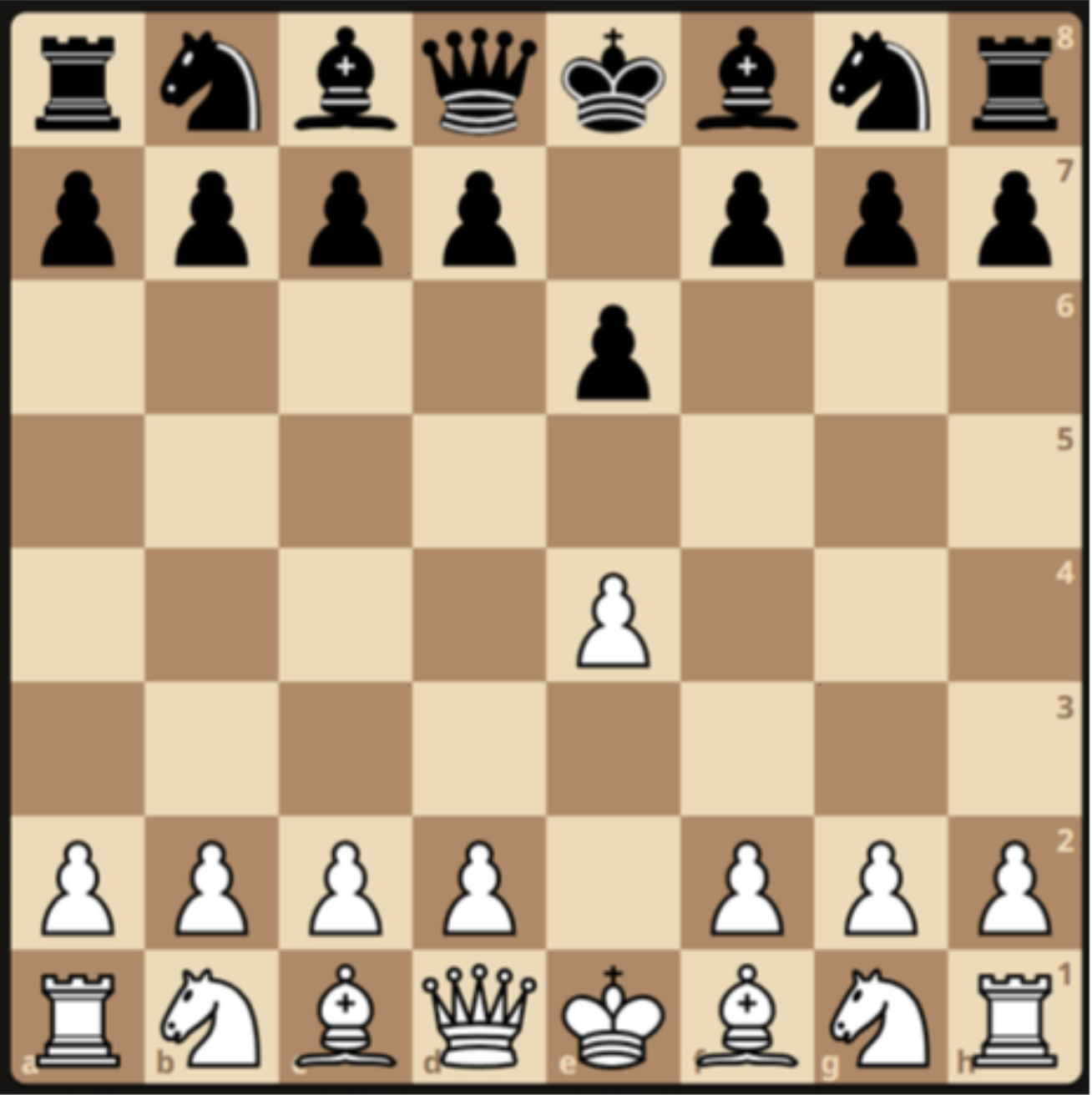}
\caption{Initial board configuration before the Ruy-Lopez opening sequence.}
\label{fig:ruy-lopez-initial}
\end{figure}

\begin{figure}[t]
\centering
\includegraphics[width=0.8\columnwidth]{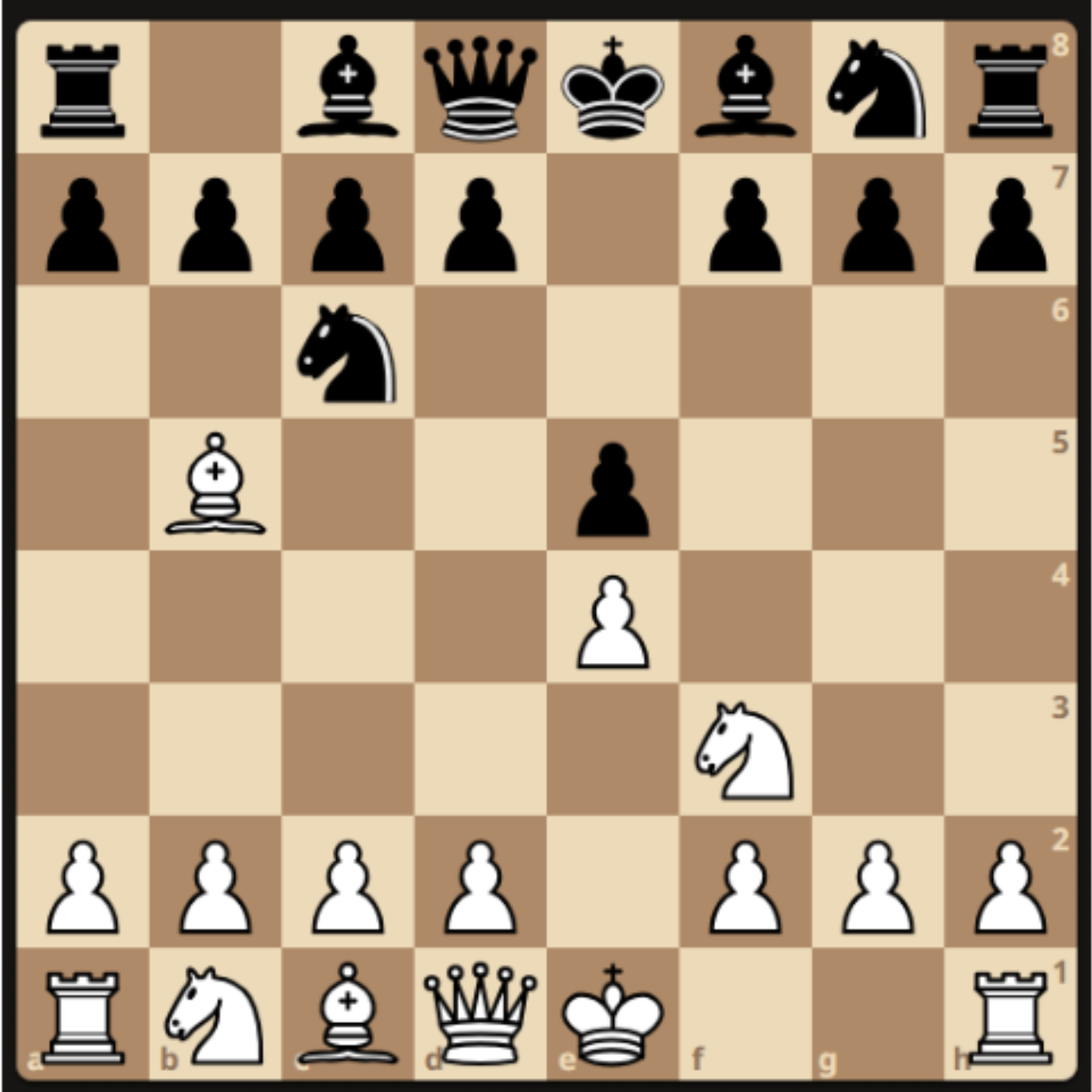}
\caption{Illustration of the Ruy-Lopez opening sequence considered in $L_g$.}
\label{fig:ruy-lopez-sequence}
\end{figure}

$L_{h} =$ {\tt
\{ You are a fruit juice manufacturing company owner with three possible
flavors: sweet, sour, and mild. Five syrups are available:
0--Orange, 1--Lemon, 2--Mango, 3--Banana, and 4--Grape. The flavor
combinations are defined as follows: Sweet = Mango + Grape;
Sour = Orange + Lemon + Grape; Mild = Banana + Mango. Construct a DFA that
accepts exactly those strings that constitute a particular flavor
according to these rules. \}
}

\end{enumerate}

Across these strategies, we include concrete instances such as maze-solving
analogies, real-world scenario encodings, and composite logical requirements
to illustrate their effect on DFA construction difficulty. Each strategy is
represented by multiple example languages in the dataset.

    \item \textbf{Mathematical Engineering Subset} (120 questions): \\
    This subset is constructed using Arden’s theorem as a core generative tool.
    Regular expressions are derived from systematically designed DFA and NFA
    transition structures, yielding symbolically precise but nonstandard DFA
    construction problems that emphasize algebraic manipulation and formal
    reasoning rather than surface familiarity.

\end{itemize}

\end{itemize}

\subsection{Data Format}

All DFA construction datasets (seen and unseen) are provided in \textbf{PDF
format}. Each problem instance includes:
\begin{itemize}
    \item The problem statement (regular expression or formal language
    specification),
    \item The corresponding DFA state transition table,
    \item A visual DFA diagram used for reference and verification.
\end{itemize}

The Knowledge Checking dataset is provided in the same PDF format and includes
both questions and explicitly labeled correct answers.

\subsection{Ground Truth and Validation}

All ground-truth DFAs and corresponding answers were manually constructed by the authors and validated through formal reasoning and automated behavioral equivalence testing, as described in the main paper. Dataset construction and validation were conducted over a period of more than three months in the context of advanced Theory of Computation coursework. The process can be summarized as follows:

\begin{itemize}[leftmargin=*]
    \item \textbf{Course Context:}
    Dataset development was carried out within two advanced Theory of Computation courses:
    \begin{itemize}
        \item \textbf{CSE 406 (Theory of Computation):} 193 undergraduate students.
        \item \textbf{CSE 525 (Advanced Theory of Computation):} 107 students, comprising both undergraduate and graduate students.
    \end{itemize}
    In total, instructional materials involved contributions across multiple course offerings from \textbf{300 students}.

    \item \textbf{Nature of Student Contributions:}
    Student contributions occurred strictly within the normal scope of coursework and assessment activities. No student-generated data was collected, analyzed, or annotated specifically for the purposes of this research, and students were not treated as research participants.

    \item \textbf{Instructional Review and Validation:}
    Instructional materials underwent systematic review and validation involving:
    \begin{itemize}
        \item \textbf{Five teaching assistants} responsible for intermediate checking, consistency verification, and instructional review.
        \item \textbf{Two faculty members} who performed final checking and validation.
        \item A dedicated \textbf{team of research assistants} supporting curation, standardization, and cross-verification.
    \end{itemize}
\end{itemize}
    All final problem instances and ground-truth solutions included in the datasets were subsequently curated, standardized, and independently verified by the authors to ensure correctness, consistency, and suitability for research use.

    For the Seen DFA Construction dataset, the repository additionally includes supporting documentation demonstrating the presence of structurally equivalent questions in publicly accessible online sources. This serves as explicit evidence of prior availability and confirms that these instances were not newly introduced for this work.

\subsection{Availability and Licensing}

All datasets, prompts, experimental scripts, validation code, evidence of
internet availability for seen questions, and representative sample outputs are
publicly released at:

\begin{center}
\url{https://anonymous.4open.science/r/dfa-llm-evaluation-B82D/}
\end{center}

The repository includes:
\begin{itemize}
    \item Dataset PDFs for all four datasets,
    \item Machine-readable metadata files describing dataset properties,
    \item Prompt templates and experimental scripts,
    \item Automated validation and evaluation code,
    \item Proof of public availability for seen dataset questions,
    \item Sample model outputs and validation reports.
\end{itemize}

All datasets are released under an explicit data license included in the
repository, permitting research use and reproducibility in accordance with conference guidelines.

\section{Appendix : Difficulty Labeling Criteria}
\label{app:EasyMed}
\subsection{Mathematical Engineering Dataset}
\label{app:regex_dataset}

For the regular-expression--based dataset, difficulty labels are assigned using a quantitative \emph{structural complexity score} derived directly from the syntax of each regular expression. The score combines the following five measurable properties:

\begin{itemize}
    \item \emph{Maximum nesting depth}, capturing hierarchical structure and long-range dependencies;
    \item \emph{Number of union operators} ($\mid$), reflecting branching and nondeterminism;
    \item \emph{Number of Kleene stars} ($*$), indicating unbounded repetition;
    \item \emph{Implicit concatenations}, representing sequential composition complexity;
    \item \emph{Expression length}, capturing overall description size (log-scaled).
\end{itemize}

These features are linearly combined into a single scalar score. Regular expressions are then partitioned into \emph{easy}, \emph{medium}, and \emph{difficult} categories using dataset-wide tertiles of this score. This procedure yields adaptive and reproducible difficulty labels without the use of manually tuned thresholds. The full implementation used in our experiments is shown below.
\begin{lstlisting}[style=codeblock, label={lst:regex_difficulty}]
  import math
  import json
  
  # -------- Regex Analysis --------
  
  def analyze_regex(regex):
      length = len(regex)
  
      union = 0
      star = 0
      concat = 0
  
      nesting = 0
      max_nesting = 0
  
      prev = None
  
      for c in regex:
          if c == '(':
              nesting += 1
              max_nesting = max(max_nesting, nesting)
          elif c == ')':
              nesting -= 1
          elif c == '|':
              union += 1
          elif c == '*':
              star += 1
          elif prev and prev not in '(|' and c not in '|)*':
              concat += 1
  
          prev = c
  
      return {
          "length": length,
          "union": union,
          "star": star,
          "concat": concat,
          "nesting": max_nesting
      }
  
  # -------- Difficulty Score --------
  
  def difficulty_score(m):
      return (
          1.5 * m["nesting"] +
          1.0 * m["union"] +
          0.5 * m["star"] +
          0.2 * m["concat"] +
          math.log1p(m["length"])
      )
  
  # -------- Dataset Classification --------
  
  def classify_dataset(entries):
      analyzed = []
  
      for e in entries:
          metrics = analyze_regex(e["regex"])
          score = difficulty_score(metrics)
          analyzed.append((e, metrics, score))
  
      scores = sorted(score for _, _, score in analyzed)
  
      q1 = scores[int(0.33 * len(scores))]
      q2 = scores[int(0.66 * len(scores))]
  
      output = []
      for e, metrics, score in analyzed:
          if score <= q1:
              difficulty = "easy"
          elif score <= q2:
              difficulty = "medium"
          else:
              difficulty = "difficult"
  
          entry = e.copy()
          entry["metrics"] = metrics
          entry["score"] = round(score, 2)
          entry["difficulty"] = difficulty
          output.append(entry)
  
      return output
  
  # -------- Example Usage --------
  
  if __name__ == "__main__":
      dataset = [
          {
              "id": "r001",
              "regex": "((b*a)|(b*a(a|b)*(a|b)))a*(a|b)a*aa*",
              "alphabet": ["a", "b"]
          }
      ]
  
      classified = classify_dataset(dataset)
      print(json.dumps(classified, indent=2))
  \end{lstlisting}

\subsection{Mathematical Art Dataset}
\label{app:math_art_dataset}

The \emph{Mathematical Art} dataset consists of natural-language DFA construction tasks characterized by multiple interacting constraints. Unlike regular-expression--based problems, these tasks are not defined by formal syntax alone; difficulty is therefore assigned based on semantic structure rather than surface form.

Specifically, difficulty labels are assigned as follows:
\begin{itemize}
    \item \textbf{Easy and medium} tasks are distinguished based on the number of explicit constraints and the extent of their interaction (e.g., positional constraints, counting constraints, and forbidden substrings).
    \item \textbf{Difficult} tasks involve narrative-based formulations, such as grid navigation, game scenarios, or real-world analogies. These problems require multi-step interpretation, implicit state tracking, and translation from informal descriptions into formal automata.
\end{itemize}

This labeling scheme reflects the additional reasoning burden imposed by narrative grounding and constraint integration, which consistently leads to higher error rates across evaluated models.

\section{Appendix: Prompt Templates}
\label{app:prompts}

This appendix provides the exact prompts used in all experiments.
Prompts are reported verbatim to ensure reproducibility.

\subsection{Prompts Used for Regular Expression to DFA Construction}

\subsubsection{Direct Input--Output Prompt}

\begin{lstlisting}
SYSTEM:
You are an expert in formal languages and automata.

USER:
Given the following regular expression and alphabet:

REGEX: {{REGEX}}
ALPHABET: {{ALPHABET}}

Task:
Construct a correct deterministic finite automaton (DFA) that recognizes exactly
the language denoted by the regular expression.

Constraints:
- The DFA must be total (every state has exactly one transition per symbol).
- Use short state names: q0, q1, q2, ...
- Do NOT include explanations, reasoning, derivations, or intermediate steps.
- Do NOT include any text outside the JSON object.

OUTPUT ONLY a single JSON object matching EXACTLY this schema:

{
  "states": ["q0", "q1", "..."],
  "alphabet": ["a", "b", "..."],
  "start_state": "q0",
  "accept_states": ["q1", "..."],
  "transitions": {
    "q0": { "a": "q1", "b": "q0" },
    "q1": { "a": "q1", "b": "q2" }
  }
}
\end{lstlisting}

\subsubsection{Chain-of-Thought (CoT) Prompt}

\begin{lstlisting}
SYSTEM:
You are an expert in formal languages and finite automata. When asked to produce a
DFA transition table, THINK step-by-step internally (Chain-of-Thought) but DO NOT
output any reasoning. Output only a single valid JSON object that exactly matches
the schema described below.

USER:
Given the following regular expression and alphabet:

REGEX: {{REGEX}}
ALPHABET: {{ALPHABET}}

Task:
1. Internally (chain-of-thought) derive a correct deterministic finite automaton
   (DFA) that recognizes the language of the regular expression.
2. Do not output any intermediate reasoning or explanation.
3. OUTPUT ONLY a single JSON object that contains the DFA transition table and
   nothing else. The JSON must match the exact schema (keys and types) below.

Required JSON schema (exact keys and types):
{
  "states": ["q0", "q1", ...],
  "alphabet": ["a", "b", ...],
  "start_state": "q0",
  "accept_states": ["q1", ...],
  "transitions": {
    "q0": { "a": "q1", "b": "q0" },
    "q1": { "a": "q1", "b": "q2" }
  }
}

Formatting rules and constraints (must follow exactly):
- The JSON object must contain only the five keys above and nothing else.
- The DFA must be total: for every state and for every symbol in "alphabet",
  there must be exactly one target state.
- Use short state names like "q0", "q1", "q2", ...
- All values must be valid JSON types.
- Do not include comments, trailing commas, or extra text.
- If construction fails, output the following machine-readable failure JSON:
  {"error":"cannot_construct","reason":"<one-
line_reason>"}

Edge-case guidance (zero-shot):
- Output ONLY the JSON object (or the failure JSON above).
- No diagnostic text before or after the JSON.
- The experiment script retries up to three times.
\end{lstlisting}

\subsubsection{Chain-of-Thought One-Shot Prompt}

\begin{lstlisting}[linewidth=\columnwidth]
SYSTEM:
You are an expert in formal languages and finite automata. When asked to produce a DFA transition table, THINK step-by-step internally (Chain-of-Thought) but DO NOT output any reasoning. Output only a single valid JSON object that exactly matches the schema described below.



USER:
Below is an example (one-shot) followed by the real input. Read the example carefully -- it demonstrates exact JSON schema, formatting, and totality rules. After the example, produce the DFA JSON for the real input only.



=== EXAMPLE (one-shot) ===
Regular expression: (a|b)*
Alphabet: ["a","b"]

Correct DFA (JSON only):
{
  "states": ["q0"],
  "alphabet": ["a", "b"],
  "start_state": "q0",
  "accept_states": ["q0"],
  "transitions": {
    "q0": { "a": "q0", "b": "q0" }
  }
}


=== END EXAMPLE ===

Now the actual input:

REGEX: {{REGEX}}
ALPHABET: {{ALPHABET}}

Task:
1. Internally (chain-of-thought) derive a correct deterministic finite automaton (DFA) that recognizes the language of the regular expression.
2. Do not output any intermediate reasoning or explanation.
3. OUTPUT ONLY a single JSON object that contains the DFA transition table and nothing else. The JSON must match the exact schema (keys and types) below.

Required JSON schema (exact keys and types):
{
  "states": ["q0", "q1", ...],
  "alphabet": ["a", "b", ...],
  "start_state": "q0",
  "accept_states": ["q1", ...],
  "transitions": {
    "q0": { "a": "q1", "b": "q0" },
    "q1": { "a": "q1", "b": "q2" }
  }
}



Formatting rules and constraints (must follow exactly):
- The JSON object must contain only the five keys above and nothing else.
- The DFA must be total: for every state and for every symbol in "alphabet", there must be exactly one target state.
- Use short state names like "q0", "q1", "q2", ... (no spaces, no special characters).
- All values must be valid JSON types (arrays, strings, objects). Do not include comments or trailing commas.
- Do not include any text before or after the JSON object (no backticks, no code fences, no extra explanation).
- If you cannot produce a correct DFA, DO NOT output {}. Instead, attempt to produce a compact but valid DFA. Only if you truly cannot construct any DFA, output a machine-readable failure JSON:
  {"error":"cannot_construct","reason":
  "<one-line_reason>"}
  (This is allowed but discouraged - prefer producing a compact DFA.)
- If the DFA is large, produce a compact encoding using short state names (q0,q1,q2,...) and include every transition for each symbol.


Edge-case guidance (one-shot):
- The above example demonstrates the EXACT output shape and formatting (including spacing/newlines are not important, but content must be valid JSON).
- The experiment script will retry up to 3 times if the first output is invalid. OUTPUT ONLY the JSON object (or the small error JSON) -- no additional text.

IMPORTANT: OUTPUT ONLY the JSON object (or the small error JSON). Do not output any other text.

\end{lstlisting}

\subsubsection{Tree-of-Thought (ToT) Direct-Branch Prompt}

\begin{lstlisting}
SYSTEM:
You are an expert in formal languages and finite automata. When asked to produce a DFA transition table, THINK step-by-step internally (Chain-of-Thought) but DO NOT output any reasoning. Output only a single valid JSON object that exactly matches the schema described below.

USER:
Given the following regular expression and alphabet:

REGEX: {{REGEX}}
ALPHABET: {{ALPHABET}}

Task:
1. Internally derive a correct deterministic finite automaton (DFA) that recognizes exactly the language denoted by the regular expression.
2. Use any sound formal reasoning internally, but do not output any intermediate steps.
3. OUTPUT ONLY a single JSON object that contains the DFA transition table and nothing else.

Required JSON schema (exact keys and types):
{
  "states": ["q0", "q1", ...],
  "alphabet": ["a", "b", ...],
  "start_state": "q0",
  "accept_states": ["q1", ...],
  "transitions": {
    "q0": { "a": "q1", "b": "q0" },
    "q1": { "a": "q1", "b": "q2" }
  }
}

Formatting rules and constraints:
- Output only the JSON object, with no extra text.
- The DFA must be total.
- Use short state names q0, q1, q2, ...
- Do not include comments, explanations, or trailing commas.
- If construction is difficult, still output a compact but valid DFA rather than an empty object.
\end{lstlisting}

\subsubsection{Tree-of-Thought (ToT) Minimization Prompt}

\begin{lstlisting}
SYSTEM:
You are an expert in deterministic finite automata. Internally aim to construct a correct DFA that recognizes the given regular expression and is close to minimal in the number of states. You may internally apply minimization techniques such as state equivalence or partition refinement. Do NOT output any reasoning.

USER:
Given the following regular expression and alphabet:

REGEX: {{REGEX}}
ALPHABET: {{ALPHABET}}

Task:
1. Internally construct a correct DFA for the regular expression.
2. Internally minimize the DFA if possible.
3. Ensure the DFA is total.
4. OUTPUT ONLY the final minimized DFA as a JSON object matching the schema below.

Required JSON schema:
{
  "states": ["q0", "q1", ...],
  "alphabet": ["a", "b", ...],
  "start_state": "q0",
  "accept_states": ["q1", ...],
  "transitions": {
    "q0": { "a": "q1", "b": "q0" },
    "q1": { "a": "q1", "b": "q2" }
  }
}

Formatting constraints:
- Output only the JSON object.
- No explanations or comments.
- DFA must be total.
\end{lstlisting}
\subsubsection{Tree-of-Thought (ToT) Derivative Method Prompt}

\begin{lstlisting}
SYSTEM:
You are an expert in formal language theory. Internally construct a DFA using regular expression derivatives (Brzozowski or Antimirov derivatives). Each DFA state corresponds to a derivative of the original regular expression. Do NOT output any intermediate reasoning.

USER:
Given the following regular expression and alphabet:

REGEX: {{REGEX}}
ALPHABET: {{ALPHABET}}

Task:
1. Internally compute the set of regular expression derivatives with respect to the alphabet.
2. Internally build the corresponding DFA from these derivatives.
3. Ensure the DFA is total.
4. OUTPUT ONLY the final DFA as a JSON object matching the exact schema below.

Required JSON schema:
{
  "states": ["q0", "q1", ...],
  "alphabet": ["a", "b", ...],
  "start_state": "q0",
  "accept_states": ["q1", ...],
  "transitions": {
    "q0": { "a": "q1", "b": "q0" },
    "q1": { "a": "q1", "b": "q2" }
  }
}

Constraints:
- Output only valid JSON.
- DFA must be total.
- No explanatory text.
- Use compact state naming.
\end{lstlisting}

\subsubsection{Tree-of-Thought (ToT) Thompson’s Construction Prompt}

\begin{lstlisting}
SYSTEM:
You are an expert in automata theory. Internally follow a Thompson-style construction: first convert the regular expression to an $\varepsilon$-NFA, then apply subset construction to obtain a DFA, and finally ensure the DFA is total. Do NOT output any intermediate structures or reasoning.



USER:
Given the following regular expression and alphabet:

REGEX: {{REGEX}}
ALPHABET: {{ALPHABET}}

Task:
1. Internally construct an $\varepsilon$-NFA using Thompson's construction.
2. Internally determinize the NFA using subset construction.
3. Internally make the DFA total by adding a sink state if necessary.
4. OUTPUT ONLY the final DFA as a single JSON object matching the schema below.

Required JSON schema:
{
  "states": ["q0", "q1", ...],
  "alphabet": ["a", "b", ...],
  "start_state": "q0",
  "accept_states": ["q1", ...],
  "transitions": {
    "q0": { "a": "q1", "b": "q0" },
    "q1": { "a": "q1", "b": "q2" }
  }
}

Formatting rules:
- Output ONLY the JSON object.
- DFA must be total.
- No explanations, comments, or extra text.
- Use short state names q0, q1, q2, ...

\end{lstlisting}

\begin{table*}
\centering
\small
\begin{tabular}{lccc}
\hline
Parameter & GPT-5.1 & Gemini-2.5 & Grok-4.1 \\
\hline
API endpoint & Public & Public & Public \\
Model variant & GPT-5.1 & Gemini-2.5-Flash & Grok-4.1-fast-reasoning \\
Temperature & 0.0 & 0.0 & 0.0 \\
Max output tokens & 4000 & 4000 & 4000 \\
Reasoning toggle & Not exposed & Not exposed & Not exposed \\
Prompt template & Identical & Identical & Identical \\
Retries & 3 & 3 & 3 \\
Timeout & 120s & 120s & 120s \\
\hline
\end{tabular}
\caption{API parameters used across model providers. No provider-specific reasoning or deliberation modes were enabled.}
\end{table*}

\subsection{Hint-Based Framework Prompts}
\label{app:hints_prompts}

This section lists the hints used to evaluate whether models can correct
their own errors under guided feedback.

\subsubsection{Initial Prompt (No Hints)}

\begin{lstlisting}[extendedchars=true,inputencoding=utf8]
Construct a deterministic finite automaton (DFA) that recognizes exactly the following language over the alphabet {a, b}:

L = (b)*(a)(a + ba)*(bb)(a + b)* + (b)*b(b)*a(a + b)*

Requirements:
- The DFA must be total.
- Use short state names (q0, q1, q2, ...).
- Specify the start state, accept states, and transitions.
- Output only the final DFA.
\end{lstlisting}

\subsubsection{Hint 1:Counterexamples protocol}

\begin{lstlisting}[extendedchars=true,inputencoding=utf8]
You are given the following language:

L = (b)*(a)(a + ba)*(bb)(a + b)* + (b)*b(b)*a(a + b)*

The DFA you previously constructed does not correctly recognize this language.

In particular, the following strings belong to L but are rejected by your DFA:
- ba
- bba
- baa

Revise or reconstruct the DFA so that it accepts all valid strings in L while continuing to reject invalid strings. Output only the corrected DFA.
\end{lstlisting}

\subsubsection{Hint 2: Error localization}

\begin{lstlisting}[extendedchars=true,inputencoding=utf8]
You are given the following language:

L = (b)*(a)(a + ba)*(bb)(a + b)* + (b)*b(b)*a(a + b)*

Your revised DFA still contains an error.

At least one transition does not correctly reflect the language definition, leading to incorrect acceptance or rejection of some strings.

Re-examine the DFA structure and transition assignments, and provide a corrected DFA that recognizes exactly L. Output only the DFA.
\end{lstlisting}

\subsubsection{Final Hint: Explicit error disclosure}

\begin{lstlisting}[extendedchars=true,inputencoding=utf8]
You are given the following language:

L = (b)*(a)(a + ba)*(bb)(a + b)* + (b)*b(b)*a(a + b)*

The remaining error arises from an incomplete interpretation of the regular expression.

In particular, the Kleene-star subexpression (a + ba)* permits arbitrary repetitions and combinations that must be fully captured by the automaton.

As a consequence of this incomplete interpretation, your DFA incorrectly accepts the string:

aabab

Reconstruct the DFA with a correct interpretation of all subexpressions and their interactions. Output only the final corrected DFA.
\end{lstlisting}

\section{Appendix: Evaluation and Validation Code}
\label{app:validation}

\paragraph{Automated Validation Pipeline.}
The automated validation pipeline provides a principled and reproducible mechanism
for assessing the correctness of LLM-generated DFAs. It combines
(i) strict structural validation, ensuring syntactic correctness, totality, and
well-formed transitions, with (ii) behavioral equivalence testing against the
ground-truth regular expression. Behavioral validation is performed using
exhaustive enumeration of all strings up to a fixed length and is supplemented
with randomized sampling of longer strings, enabling detection of both local and
global acceptance errors. This design ensures that reported correctness reflects
semantic language equivalence rather than surface-level plausibility.

The validator is model-agnostic and deterministic, producing identical results
given the same DFA and regular expression. It records explicit counterexamples
whenever a mismatch is detected, supporting transparent error analysis and
independent verification. All validation artifacts, including per-instance
reports and aggregated summaries, are saved to disk to facilitate reproducibility
and post-hoc inspection.

At the same time, the validation procedure has inherent limitations. Exhaustive
testing is bounded by a maximum string length, and while random sampling extends
coverage, formal equivalence between a DFA and a regular expression is undecidable
via finite testing alone. As a result, DFAs classified as correct should be
interpreted as \emph{likely correct} within the tested bounds rather than
formally proven equivalent. Additionally, the validator assumes the correctness
of the regular expression semantics as interpreted by the host regex engine,
which may differ from theoretical automata semantics in edge cases. These
limitations are mitigated in the main evaluation by complementary human expert
inspection, but they remain important considerations when interpreting automated
results.

\subsection{Automated DFA Validation Pipeline}
\label{app:validator}

\begin{lstlisting}
#!/usr/bin/env python3
"""
validate_dfa_outputs.py

Automated DFA validation pipeline used in experiments.

Validation protocol:
1. Schema validation (JSON structure + required fields).
2. Totality validation (every state has one transition per symbol).
3. Behavioral equivalence testing against the ground-truth regular expression:
   - Exhaustive enumeration up to MAX_EXHAUSTIVE_LEN.
   - Large-scale randomized testing for longer strings.

Outputs:
- Per-DFA detailed validation reports (JSON).
- Aggregated summary table (CSV + JSON).

This script is model-agnostic and dataset-agnostic.
"""

import os
import json
import re
import random
import csv
from itertools import product
from typing import List, Dict, Tuple

# =============================
# CONFIGURATION (EDIT HERE)
# =============================

TABLES_DIR = "outputs/tables"
RAW_DIR = "outputs/raw"
VALID_DIR = "outputs/validation"

MAX_EXHAUSTIVE_LEN = 6        # exhaustive testing for strings of length 0..6
N_RANDOM = 2000               # number of random test strings
MAX_RANDOM_LEN = 15           # maximum length of random strings
MAX_COUNTEREXAMPLES = 100     # cap stored counterexamples per DFA
RANDOM_SEED = 42              # reproducibility

# =============================
# INITIALIZATION
# =============================

os.makedirs(VALID_DIR, exist_ok=True)
random.seed(RANDOM_SEED)

# =============================
# UTILITIES
# =============================

def load_json(path: str):
    with open(path, "r", encoding="utf-8") as f:
        return json.load(f)

def save_json(path: str, obj):
    with open(path, "w", encoding="utf-8") as f:
        json.dump(obj, f, indent=2, ensure_ascii=False)

def is_valid_schema(dfa: Dict) -> Tuple[bool, str]:
    required = {"states", "alphabet", "start_state", "accept_states", "transitions"}
    if not isinstance(dfa, dict):
        return False, "not_json_object"
    if not required.issubset(dfa.keys()):
        return False, "missing_required_keys"
    if not isinstance(dfa["states"], list):
        return False, "states_not_list"
    if not isinstance(dfa["alphabet"], list):
        return False, "alphabet_not_list"
    if dfa["start_state"] not in dfa["states"]:
        return False, "invalid_start_state"
    if not isinstance(dfa["accept_states"], list):
        return False, "accept_states_not_list"
    if not isinstance(dfa["transitions"], dict):
        return False, "transitions_not_dict"
    return True, "ok"

def check_totality(dfa: Dict):
    missing = []
    invalid_targets = []

    for s in dfa["states"]:
        if s not in dfa["transitions"]:
            missing.append({"state": s, "reason": "no_transition_block"})
            continue
        for a in dfa["alphabet"]:
            if a not in dfa["transitions"][s]:
                missing.append({"state": s, "symbol": a})
            else:
                tgt = dfa["transitions"][s][a]
                if tgt not in dfa["states"]:
                    invalid_targets.append({"state": s, "symbol": a, "target": tgt})

    return missing, invalid_targets

def simulate_dfa(dfa: Dict, word: str) -> bool:
    cur = dfa["start_state"]
    for ch in word:
        if ch not in dfa["alphabet"]:
            return False
        cur = dfa["transitions"].get(cur, {}).get(ch)
        if cur is None:
            return False
    return cur in dfa["accept_states"]

def generate_exhaustive_strings(alphabet: List[str], max_len: int) -> List[str]:
    strings = [""]
    for L in range(1, max_len + 1):
        for tup in product(alphabet, repeat=L):
            strings.append("".join(tup))
    return strings

def generate_random_strings(alphabet: List[str], n: int, max_len: int) -> List[str]:
    out = set()
    while len(out) < n:
        L = random.randint(0, max_len)
        out.add("".join(random.choice(alphabet) for _ in range(L)))
    return list(out)

# =============================
# MAIN VALIDATION LOOP
# =============================

def main():
    table_files = [
        os.path.join(TABLES_DIR, f)
        for f in os.listdir(TABLES_DIR)
        if f.endswith(".json")
    ]

    summary = []

    for table_path in table_files:
        base = os.path.splitext(os.path.basename
        (table_path))[0]
        print(f"[VALIDATING] {base}")

        report = {
            "id": base,
            "schema_valid": False,
            "schema_error": None,
            "totality_missing": [],
            "totality_invalid_targets": [],
            "tested_strings": 0,
            "counterexamples": [],
            "verdict": ""
        }

        try:
            dfa = load_json(table_path)
        except Exception as e:
            report["schema_error"] = f"json_load_error: {e}"
            report["verdict"] = "invalid_json"
            save_json(os.path.join(VALID_DIR, base + "_report.json"), report)
            continue

        ok, reason = is_valid_schema(dfa)
        report["schema_valid"] = ok
        report["schema_error"] = None if ok else reason

        if not ok:
            report["verdict"] = "schema_invalid"
            save_json(os.path.join(VALID_DIR, base + "_report.json"), report)
            continue

        missing, invalid = check_totality(dfa)
        report["totality_missing"] = missing
        report["totality_invalid_targets"] = invalid

        # Locate matching raw file for regex
        regex = None
        for rf in os.listdir(RAW_DIR):
            if rf.startswith(base):
                raw = load_json(os.path.join(RAW_DIR, rf))
                regex = raw.get("regex")
                break

        if regex is None:
            report["verdict"] = "no_ground_truth_regex"
            save_json(os.path.join(VALID_DIR, base + "_report.json"), report)
            continue

        try:
            pattern = re.compile(f"^({regex})$")
        except Exception as e:
            report["verdict"] = f"regex_compile_error: {e}"
            save_json(os.path.join(VALID_DIR, base + "_report.json"), report)
            continue

        exhaustive = generate_exhaustive_strings
        (dfa["alphabet"], MAX_EXHAUSTIVE_LEN)
        random_tests = generate_random_strings(dfa["alphabet"], N_RANDOM, MAX_RANDOM_LEN)
        tests = exhaustive + [s for s in random_tests if s not in exhaustive]

        for w in tests:
            dfa_accept = simulate_dfa(dfa, w)
            regex_accept = pattern.fullmatch(w) is not None
            report["tested_strings"] += 1

            if dfa_accept != regex_accept:
                report["counterexamples"].append({
                    "string": w,
                    "dfa_accepts": dfa_accept,
                    "regex_accepts": regex_accept
                })
                if len(report["counterexamples"]) >= MAX_COUNTEREXAMPLES:
                    break

        if missing or invalid:
            report["verdict"] = "not_total"
        elif report["counterexamples"]:
            report["verdict"] = "incorrect"
        else:
            report["verdict"] = "likely_correct"

        save_json(os.path.join(VALID_DIR, base + "_report.json"), report)

        summary.append({
            "id": base,
            "schema_valid": ok,
            "totality_ok": not (missing or invalid),
            "tested_strings": report["tested_strings"],
            "num_counterexamples": len(report["counterexamples"]),
            "verdict": report["verdict"]
        })

    # =============================
    # WRITE SUMMARY TABLES
    # =============================

    with open(os.path.join(VALID_DIR, "summary.csv"), "w", newline="", encoding="utf-8") as f:
        writer = csv.DictWriter(
            f,
            fieldnames=["id", "schema_valid", "totality_ok",
                        "tested_strings", "num_counterexamples", "verdict"]
        )
        writer.writeheader()
        for row in summary:
            writer.writerow(row)

    save_json(os.path.join(VALID_DIR, "summary.json"), summary)

    print("[DONE] DFA validation completed.")

if __name__ == "__main__":
    main()
\end{lstlisting}

\section{Appendix: Experiment Execution Code}
\label{app:execution_code}
\paragraph{Experiment Execution Code.}
We provide the full experiment execution code used to query language models and
collect DFA outputs. The script is model-agnostic and was used uniformly across
all evaluated models and prompting strategies, differing only in the model
identifier and prompt templates. It enforces deterministic decoding, robust
retry logic, strict JSON extraction, and complete logging of raw outputs, parsed
DFAs, and metadata. All runs write outputs regardless of success or failure,
enabling reproducible analysis of both correctness and failure modes.

\paragraph{Strengths and Limitations.}
The execution pipeline ensures reproducibility through deterministic decoding
and uniform evaluation settings, and robustness through retries and structured
output validation. However, it does not attempt to correct or post-process
invalid DFAs, and relies on downstream automated and human validation for
semantic correctness. Timeout events and malformed outputs are treated as
unsuccessful attempts, reflecting practical deployment constraints rather than
purely theoretical capability.

\subsection{Experiment Execution Code}
\label{app:execution_subsection}

\begin{lstlisting}
#!/usr/bin/env python3
"""
run_experiment_tot.py

General Tree-of-Thought (ToT) experiment runner for LLM-based DFA construction.

Features:
- Model-agnostic (OpenAI / API-based LLMs)
- Deterministic decoding (temperature = 0)
- Robust JSON extraction and retry logic
- Strict output logging (raw, parsed, metadata)
- Always writes outputs (success or failure)
- Reproducible and configurable via constants

This script is used uniformly across models and prompting strategies by
changing only the MODEL identifier and prompt templates.
"""

import os
import json
import time
import re
from datetime import datetime
from typing import Optional
from openai import OpenAI, OpenAIError

# =============================
# CONFIGURATION (MODEL-AGNOSTIC)
# =============================

MODEL = os.environ.get("LLM_MODEL", "gpt-5.1")
TEMPERATURE = 0.0
MAX_TOKENS = 4000

DATASET_PATH = "data/analysis_data.json"

PROMPT_FILES = {
    "direct":     "prompts/tot_direct.txt",
    "minimal":    "prompts/tot_minimal.txt",
    "derivative": "prompts/tot_derivative.txt",
    "thompson":   "prompts/tot_thompson.txt",
}

OUTPUT_ROOT = "outputs/tot_experiments"

RETRIES = 3
RETRY_SLEEP = 1.5
RATE_LIMIT_SLEEP = 0.3

# =============================
# INIT
# =============================

client = OpenAI()

TIMESTAMP = datetime.utcnow().strftime("%Y%m%dT%H%M%
SZ")
OUTPUT_RAW_DIR   = os.path.join(OUTPUT_ROOT, MODEL, "raw")
OUTPUT_TABLE_DIR = os.path.join(OUTPUT_ROOT, MODEL, "tables")
OUTPUT_META_DIR  = os.path.join(OUTPUT_ROOT, MODEL, "meta")

for d in [OUTPUT_RAW_DIR, OUTPUT_TABLE_DIR, OUTPUT_META_DIR]:
    os.makedirs(d, exist_ok=True)

PROJECT_ROOT = os.path.abspath(os.path.join(os.path.
dirname(__file__), ".."))

# =============================
# UTILITIES
# =============================

def now_ts():
    return datetime.utcnow().strftime("%Y%m%dT%H%M
%SZ")

def safe_filename(s: str) -> str:
    return re.sub(r"[^0-9A-Za-z._-]", "_", s)[:200]

def load_json(path):
    with open(path, "r", encoding="utf-8") as f:
        return json.load(f)

def load_prompt(path):
    with open(path, "r", encoding="utf-8") as f:
        return f.read()

def save_json(path, obj):
    with open(path, "w", encoding="utf-8") as f:
        json.dump(obj, f, indent=2, ensure_ascii=False)

def try_parse_json(s: str) -> Optional[dict]:
    if not s:
        return None
    s = re.sub(r"```(?:json)?", "", s, flags=re.IGNORECASE).strip("` \n")
    try:
        return json.loads(s)
    except Exception:
        pass
    m = re.search(r"\{[\s\S]*\}", s)
    if m:
        try:
            return json.loads(m.group(0))
        except Exception:
            pass
    return None

# =============================
# TOKEN DISPATCH
# =============================

def completion_kwargs():
    if MODEL.startswith("gpt-5"):
        return {"max_completion_tokens": MAX_TOKENS}
    return {"max_tokens": MAX_TOKENS}

# =============================
# MODEL ACCESS CHECK
# =============================

def check_model_access():
    try:
        r = client.chat.completions.create(
            model=MODEL,
            messages=[{"role": "user", "content": "ping"}],
            temperature=0.0,
            **completion_kwargs()
        )
        actual = getattr(r, "model", None)
        if actual is None or not actual.startswith(MODEL.split("-")[0]):
            raise RuntimeError(f"Model mismatch: requested={MODEL}, actual={actual}")
    except OpenAIError as e:
        raise RuntimeError(f"Model access failed: {e}")

# =============================
# PROMPTING
# =============================

def build_messages(template, regex, alphabet):
    filled = (
        template
        .replace("{{REGEX}}", regex)
        .replace("{{ALPHABET}}", json.dumps(alphabet))
    )
    return [
        {
            "role": "system",
            "content": (
                "You are an expert in formal languages and automata. "
                "Internally reason as needed, but OUTPUT ONLY the DFA JSON."
            )
        },
        {"role": "user", "content": filled}
    ]

def call_model(messages):
    r = client.chat.completions.create(
        model=MODEL,
        messages=messages,
        temperature=TEMPERATURE,
        **completion_kwargs()
    )
    text = r.choices[0].message.content or ""
    meta = {
        "requested_model": MODEL,
        "actual_model": getattr(r, "model", None),
        "finish_reason": r.choices[0].finish_reason,
        "usage": getattr(r, "usage", None)
    }
    return text, meta

# =============================
# CORE LOOP
# =============================

def run_single_branch(entry, branch, prompt_template):
    regex_id = entry["id"]
    regex = entry["regex"]
    alphabet = entry.get("alphabet", [])
    ts = now_ts()

    messages = build_messages(prompt_template, regex, alphabet)

    raw_text, meta, parsed = "", None, None
    attempts = 0

    while attempts < RETRIES:
        attempts += 1
        try:
            raw_text, meta = call_model(messages)
        except Exception as e:
            meta = {"error": str(e)}
            time.sleep(RETRY_SLEEP)
            continue

        parsed = try_parse_json(raw_text)
        if parsed:
            break

        messages.append({
            "role": "user",
            "content": "OUTPUT ONLY the DFA JSON. No explanations."
        })
        time.sleep(RETRY_SLEEP)

    base = safe_filename(f"{regex_id}_{branch}_{ts}")

    save_json(os.path.join(OUTPUT_RAW_DIR, base + ".json"), {
        "id": regex_id,
        "branch": branch,
        "regex": regex,
        "alphabet": alphabet,
        "raw_output": raw_text,
        "attempts": attempts,
        "meta": meta,
        "model": MODEL,
        "timestamp": ts
    })

    save_json(os.path.join(OUTPUT_META_DIR, base + "_meta.json"), meta)

    table_obj = parsed if parsed else {
        "error": "GENERATION_FAILED",
        "reason": "Invalid or non-JSON output",
        "regex_id": regex_id,
        "branch": branch,
        "model": MODEL,
        "attempts": attempts
    }

    save_json(os.path.join(OUTPUT_TABLE_DIR, base + ".json"), table_obj)

# =============================
# MAIN
# =============================

def main():
    check_model_access()
    dataset = load_json(os.path.join(PROJECT_ROOT, DATASET_PATH))
    prompts = {
        k: load_prompt(os.path.join(PROJECT_ROOT, v))
        for k, v in PROMPT_FILES.items()
    }

    for entry in dataset:
        for branch, tmpl in prompts.items():
            run_single_branch(entry, branch, tmpl)
            time.sleep(RATE_LIMIT_SLEEP)

if __name__ == "__main__":
    main()
\end{lstlisting}

\section{Appendix: Common Problems Identified from LLM-Generated DFAs}
\label{app:failure-modes}

This appendix summarizes recurring error patterns observed in deterministic
finite automaton (DFA) constructions produced by large language models (LLMs)
across different prompting strategies. Each problem type is illustrated using a
concrete example and a corresponding DFA diagram. The goal is diagnostic: to
characterize systematic construction failures rather than isolated errors.

\subsection*{Summary of Identified Problem Types}

\begin{table*}[t]
\centering
\caption{Summary of Common DFA Construction Problems}
\label{tab:problem-summary}
\small
\begin{tabular}{p{0.25\textwidth} p{0.65\textwidth}}
\toprule
\textbf{Problem} & \textbf{Identified Issue} \\
\midrule
Problem 1 & Disregard for Kleene star semantics \\
Problem 2 & Violation of Brzozowski derivative semantics \\
Problem 3 & Errors in initial DFA construction before minimization \\
Problem 4 & Over-acceptance of strings outside the target language \\
Problem 5 & Failure to preserve constraints under concatenation \\
Problem 6 & Introduction of redundant or unreachable states \\
\bottomrule
\end{tabular}
\end{table*}

\subsection*{Problem 1: Disregard for Kleene Star Semantics}

\begin{figure}[t]
\centering
\includegraphics[width=0.7\columnwidth]{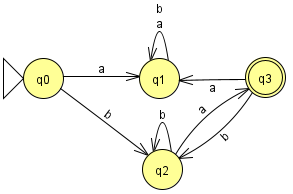}
\caption{DFA generated via direct construction for $L = b(a \mid b)^{*}ab$.}
\label{fig:problem1}
\end{figure}

\noindent
Figure~\ref{fig:problem1} shows a DFA generated via direct construction for the
language $L = b(a \mid b)^{*}ab$.
The model correctly identifies the terminal pattern \texttt{ab} but fails to
account for the Kleene star $(a \mid b)^{*}$, resulting in a DFA that accepts
strings ending in \texttt{ba} rather than enforcing the required suffix
\texttt{ab}. This error reflects an incomplete semantic interpretation of
unbounded repetition under the Kleene star.

\subsection*{Problem 2: Violation of Brzozowski Derivative Semantics}

\begin{figure}[t]
\centering
\includegraphics[width=0.7\columnwidth]{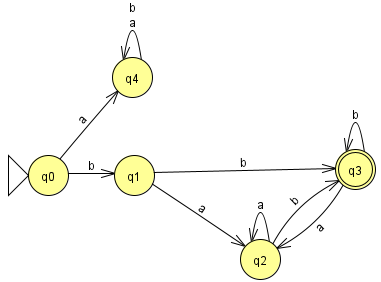}
\caption{Derivative-based DFA generated for $L = b(a \mid b)^{*}ab$.}
\label{fig:problem2}
\end{figure}

\noindent\textbf{Construction method:} Derivative-based construction.
\noindent
Figure~\ref{fig:problem2} shows a derivative-based DFA generated for the language
$L = b(a \mid b)^{*}ab$.
The construction fails to normalize semantically equivalent Brzozowski
derivatives and incorrectly reintroduces consumed prefixes. As a result,
distinct residual languages are treated as separate states, violating the
formal semantics of regular expression derivatives and yielding an incorrect
DFA.

\subsection*{Problem 3: Errors in Initial DFA Construction}

\begin{figure}[t]
\centering
\includegraphics[width=0.7\columnwidth]{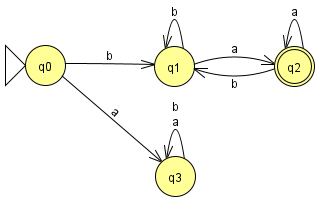}
\caption{DFA produced prior to minimization for $L = b(a \mid b)^{*}ab$.}
\label{fig:problem3}
\end{figure}

\noindent\textbf{Construction method:} Minimization-based construction.
\noindent
Figure~\ref{fig:problem3} shows a DFA produced prior to minimization for the
language $L = b(a \mid b)^{*}ab$.
The initial DFA is incorrectly constructed due to faulty state semantics and
transition assignments. Subsequent minimization amplifies these errors by
merging states that should remain distinct, resulting in an invalid minimized
DFA.

\subsection*{Problem 4: Over-Acceptance of Strings}

\begin{figure}[t]
\centering
\includegraphics[width=0.7\columnwidth]{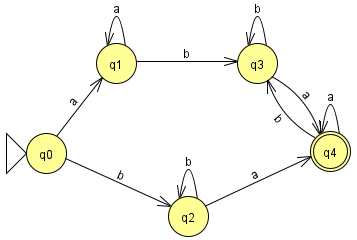}
\caption{Over-accepting DFA generated for the target language.}
\label{fig:problem4}
\end{figure}

\noindent\textbf{Construction method:} Derivative-based construction.
\noindent
Figure~\ref{fig:problem4} shows an over-accepting DFA generated for the language
\[
\begin{aligned}
L = {} & ((a \mid b)^{*}  (aa^{*}bb^{*} \mid b(b \mid ab)^{*}aaa^{*}bb^{*}) \\
       & (aa^{*}bb^{*})^{*} \\
       & \mid (a \mid b)^{*}b(b \mid ab)^{*}a)aa^{*}.
\end{aligned}
\]

The constructed DFA accepts strings that do not belong to the target language:
formally, there exists a string $w \in \Sigma^{*}$ such that
$w \in L(\mathrm{DFA})$ while $w \notin L$. This over-acceptance arises when
syntactically similar but semantically distinct derivatives are merged without
establishing true language equivalence. In this example, strings of the form
$bb^{*}a$ are incorrectly accepted.

\subsection*{Problem 5: Failure to Preserve Constraints under Concatenation}

\begin{figure}[t]
\centering
\includegraphics[width=0.7\columnwidth]{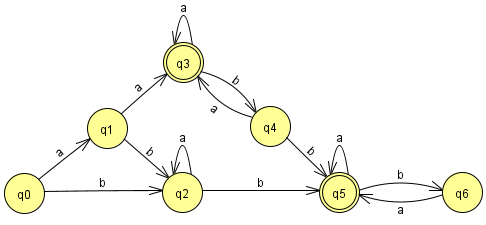}
\caption{DFA failing to preserve constraints under concatenation.}
\label{fig:problem5}
\end{figure}

\noindent\textbf{Construction methods:} Direct and Thompson constructions.
\noindent
Figure~\ref{fig:problem5} shows a DFA that fails to preserve constraints under
concatenation for the language
\[
L = ((a^{*}ba^{*}b \mid \allowbreak
(a^{*}ba^{*}b \mid a^{*}a)\allowbreak
(b \mid a)^{*}a)\allowbreak
)a^{*}.
\]
The constructed DFA correctly enforces individual sub-constraints such as
$a^{*}ba^{*}b$ and $a^{*}a$ in isolation. However, when these constraints are
composed via concatenation, global boundary conditions are not preserved.

\subsection*{Problem 6: Introduction of Redundant or Unreachable States}

\noindent\textbf{Construction methods:} Direct, Thompson, derivative-based, and
minimization-based constructions.

\begin{figure}[t]
\centering
\includegraphics[width=0.7\columnwidth]{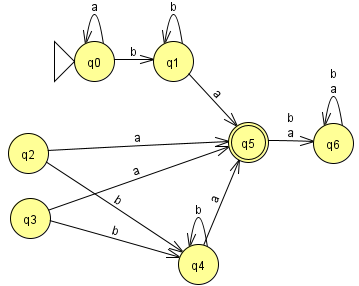}
\caption{DFA containing redundant and unreachable states.}
\label{fig:problem6}
\end{figure}

\noindent
Figure~\ref{fig:problem6} shows a DFA containing redundant and unreachable states
for the language
\[
\begin{aligned}
L = {} & ((a^{*}b \mid (a^{*}a)a^{*}b \mid (a^{*}b)b^{*}b \\
       & \mid ((a)^{*}ba \mid (a^{*}a)a^{*}ba \mid (a^{*}b)b^{*}ba) \\
       & (aa \mid b)^{*}a)(a \mid b) \mid (a^{*}b)b^{*})a.
\end{aligned}
\]

The constructed DFA contains multiple states that are unreachable from the start
state and therefore do not contribute to language recognition. In some examples,
some of these states are accepting despite having no incoming paths. This
indicates that the model introduces states corresponding to syntactic components
of the regular expression without performing reachability or usefulness
analysis, resulting in redundant and semantically irrelevant states.

\begin{figure*}
\begin{center}
\begin{tabular}{c}
\includegraphics[width=5.0in]{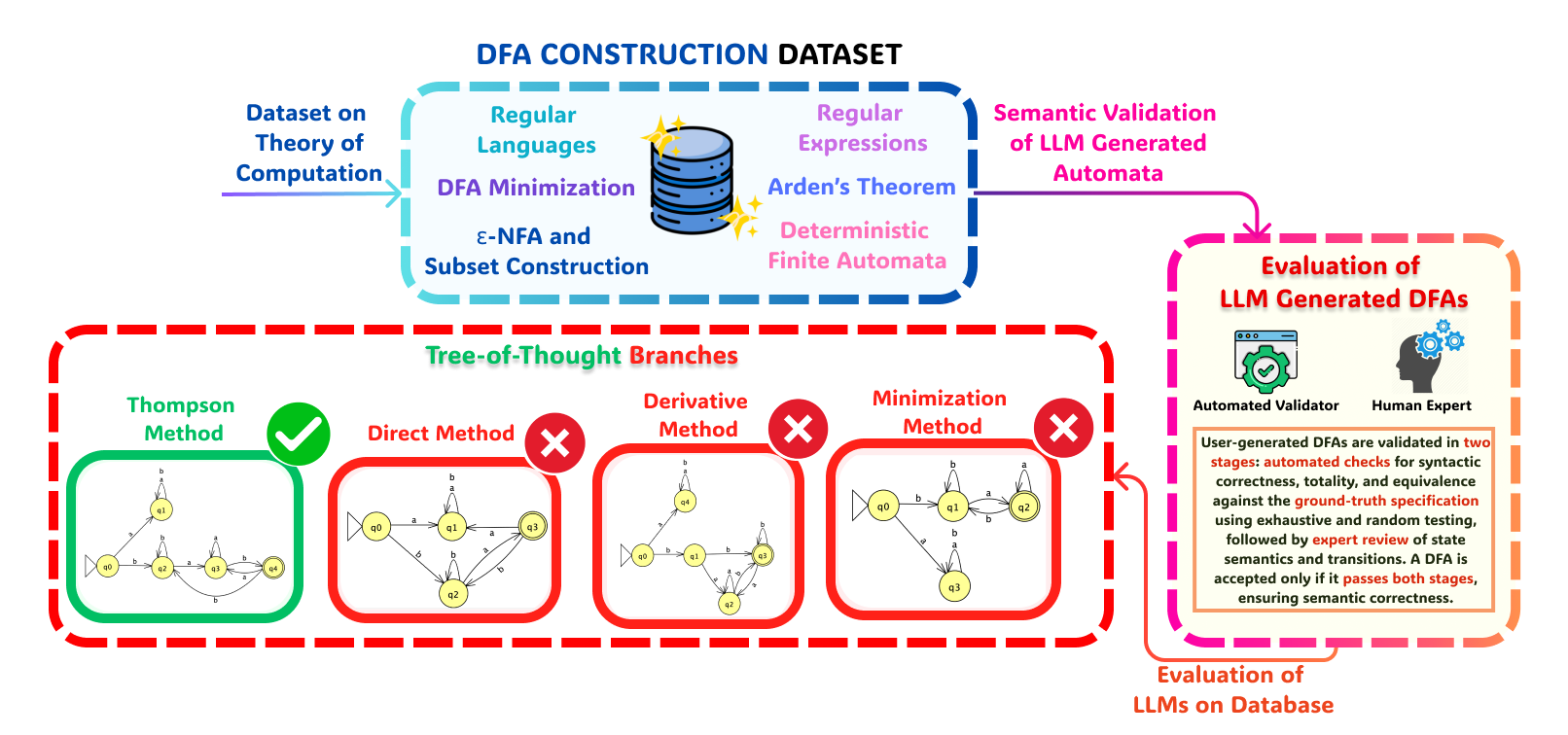} 
\end{tabular}
\caption{DFA construction dataset and evaluation pipeline, with representative DFA outputs generated by ToT prompting following construction strategies for the language $L_9 = b(a+b)^*ab$.}
\label{FigL4}
\end{center}
\end{figure*}

For an overall view, Figure \ref{FigL4} shows the common mistakes for the Language $L_9$ under the ToT methods.

\section{Appendix : Derivative-Based DFA Output for $L_9$}
\label{app:derivative_L8}

This appendix reports the exact output produced by the derivative-based DFA
construction for language $L_9$. The content is shown verbatim as returned by
the model, without normalization, correction, or post-processing.

\subsubsection{Model Output (Derivative Construction for $L_9$)}
\begin{lstlisting}[language=json,extendedchars=true,inputencoding=utf8]
  {
    "method": "derivative",
    "derivatives": {
      "q0": {
        "regex": "b(a|b)*ab",
        "nullable": false
      },
      "q1": {
        "regex": "(a|b)*ab",
        "nullable": false
      },
      "q2": {
        "regex": "b(a|b)*ab+(a|b)*b",
        "nullable": false
      },
      "q3": {
        "regex": "(a|b)*ab+(a|b)*",
        "nullable": true
      },
      "q4": {
        "regex": "\phi",
        "nullable": false
      }
    },
    "dfa": {
      "states": [
        "q0",
        "q1",
        "q2",
        "q3",
        "q4"
      ],
      "alphabet": [
        "a",
        "b"
      ],
      "start_state": "q0",
      "accept_states": [
        "q3"
      ],
      "transitions": {
        "q0": {
          "a": "q4",
          "b": "q1"
        },
        "q1": {
          "a": "q2",
          "b": "q3"
        },
        "q2": {
          "a": "q2",
          "b": "q3"
        },
        "q3": {
          "a": "q2",
          "b": "q3"
        },
        "q4": {
          "a": "q4",
          "b": "q4"
        }
      }
    }
  }
  
\end{lstlisting}

\section{Appendix: Determinism and Randomness Control}
\label{app:determinism}

All experiments were designed to be as deterministic and reproducible as possible, subject to the constraints of black-box LLM APIs.

\paragraph{Decoding determinism.}
For all models and prompting strategies, decoding temperature was fixed to $0.0$, and no stochastic sampling parameters were modified. This ensures that, for a given prompt and model endpoint, the generation process is deterministic to the extent supported by the provider.

\paragraph{Prompt determinism.}
Each prompt template is fixed and version-controlled in the repository. For Tree-of-Thought and Chain-of-Thought settings, a fixed, strategy-specific
prompt template is reused across all runs. No adaptive prompt modification or dynamic hint generation is performed during a single model invocation. 

\paragraph{Retry policy.}
To mitigate transient API failures and formatting errors, each query is retried up to a fixed number of attempts using an identical prompt. Retries are only triggered when the output is invalid (e.g., non-JSON or schema-violating) and do not introduce additional randomness or alternative prompts.

\paragraph{Validation randomness control.}
Behavioral equivalence testing uses a hybrid validation strategy consisting of:
(i) exhaustive enumeration of all strings up to a fixed maximum length, and
(ii) randomized testing over longer strings.
All randomized testing is performed with a fixed random seed, ensuring that validation results are fully reproducible.

\paragraph{Non-determinism across API calls.}
Despite the above controls, repeated API calls to the same model may still produce different outputs due to undocumented provider-side nondeterminism (e.g., model updates, inference-time optimizations, or distributed serving). Such variation is treated as an inherent property of deployed LLM systems rather than experimental noise.

\paragraph{Reporting policy.}
All reported results correspond to the actual outputs returned by the API during evaluation and are not post-selected. Timeouts, formatting failures, and invalid outputs are logged explicitly and counted according to the evaluation protocol described in the main paper.

\section{Appendix: Runtime Environment}
\label{app:runtime}

All experiments were executed using custom Python scripts on a local workstation.
The implementation was written in Python (version 3.10+) and relied exclusively
on standard libraries and official model APIs. No proprietary or undocumented
tooling was used.

All LLM interactions were performed via API calls using deterministic decoding
settings (temperature set to $0$). Experiments were conducted sequentially with
explicit rate-limiting and retry logic to ensure stability and reproducibility.
Each model invocation was stateless: no conversational context was carried across
calls. This includes the hint-based framework, where the full problem
specification was re-provided at every stage.

Model outputs, metadata (including token usage and finish reasons), parsed DFA
tables, and validation reports were logged to disk in structured JSON format. The
runtime environment enforced strict output handling: results were recorded for
every attempt, including malformed outputs, timeouts, or generation failures.

Validation and evaluation were performed offline using a separate automated
pipeline that exhaustively and randomly tested DFA behavior against the
ground-truth regular expressions. All randomness used in validation (e.g., random
string generation) was controlled via fixed random seeds.

No fine-tuning, model-side configuration changes, or system-level optimizations
were applied. Differences in performance therefore reflect intrinsic model
behavior under identical runtime conditions rather than environmental
variability.

\end{document}